\documentclass[10pt,twocolumn,letterpaper]{article}

\usepackage{wacv}
\usepackage{times}
\usepackage{epsfig}
\usepackage{graphicx}
\usepackage{amsmath}
\usepackage{amssymb}
\usepackage[accsupp]{axessibility}  

\usepackage{xcolor}
\usepackage{booktabs}

\usepackage[utf8x]{inputenc} 

\usepackage{calc}

\usepackage[numbers,sort]{natbib}

\usepackage{dblfloatfix}

\usepackage{array, multirow}
\newcolumntype{C}[1]{>{\centering\let\newline\\\arraybackslash\hspace{0pt}}m{#1}}
\newcolumntype{L}{>{$}l<{$}}

\newcommand{\NA}{-}

\newcommand\vertarrowbox[3][6ex]{%
\begin{array}[t]{@{}c@{}} #2 \\
\left\uparrow\vcenter{\hrule height #1}\right.\kern-\nulldelimiterspace\\
\makebox[0pt]{\scriptsize#3}
\end{array}%
}

\newcommand{\bb}[1]{#1}

\newcommand{\bx}{\bb{x}}

\newcommand{\by}{\bb{y}}

\newcommand{\bh}{\bb{h}}

\newcommand{\bz}{\bb{z}}

\renewcommand{\boldsymbol}[1]{#1}
\newcommand{\bT}{{\boldsymbol{\theta}}}

\newcommand{\pz}{p_{\bz}}
\newcommand{\pyGivenx}{p_{\by|\bx}}

\newcommand{\fT}{f_{\bT}}

\newcommand{\parsection}[1]{\vspace{0.5mm}\noindent\textbf{#1:}~}

\newcommand{\reals}{\mathbb{R}}

\def\wacvPaperID{608} 

\wacvfinalcopy 

\ifwacvfinal
\def\assignedStartPage{9876} 
\fi

\ifwacvfinal
\usepackage[breaklinks=true,bookmarks=false]{hyperref}
\else
\usepackage[pagebackref=true,breaklinks=true,colorlinks,bookmarks=false]{hyperref}
\fi

\ifwacvfinal
\else
\fi

\usepackage{capt-of,etoolbox}
\makeatletter
\apptocmd\@maketitle{{\introfig{}\par}}{}{}
\makeatother

\begin{document} 

\title{Normalizing Flow as a Flexible Fidelity Objective for Photo-Realistic Super-resolution}

\newcommand{\aand}{\hspace{6mm}}
\author{Andreas Lugmayr \aand Martin Danelljan \aand Fisher Yu \aand Luc Van Gool \aand Radu Timofte \vspace{1.5mm}\\
	CVL, ETH Z\"urich, Switzerland
}

\newcommand{\introfig}{\protect\centering\vspace{-5mm}
\includegraphics[width=0.98\linewidth]{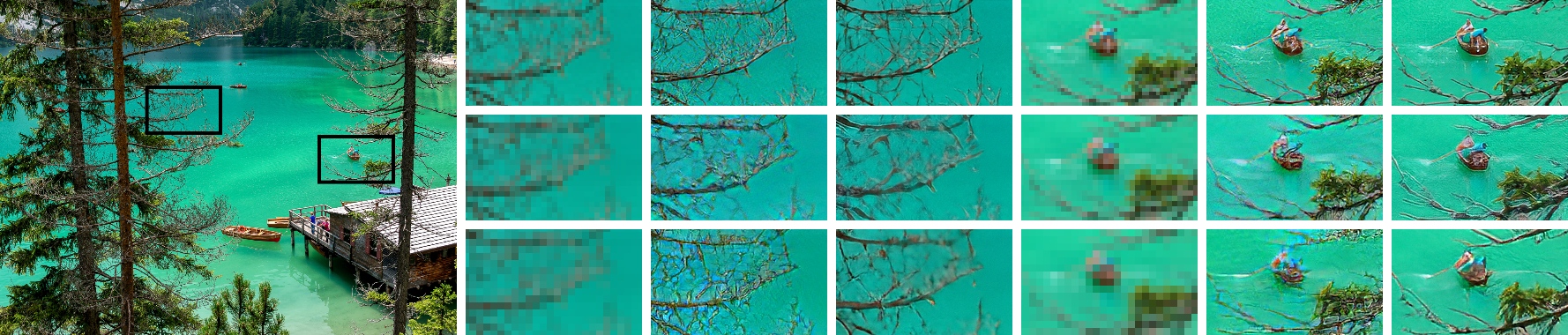}~%
\rotatebox[origin=l]{90}{\resizebox{3.6cm}{!}{
		\begin{tabular}{ C{1.2cm} C{1.2cm} C{1.2cm}}
			8$\times$ & 6$\times$ & 4$\times$
		\end{tabular}
}}\vspace{-2mm}
\resizebox{1.00\linewidth}{!}{%
\begin{tabular}{ C{8cm} C{3cm} C{3cm} C{3cm} C{3cm} C{3cm} C{3cm} C{0.2cm} }
& Low-Resolution & ESRGAN~\cite{wang2018esrgan} & \textbf{Ours} & Low-Resolution & ESRGAN~\cite{wang2018esrgan} & \textbf{Ours} &
\end{tabular}}\vspace{0mm}
  \captionof{figure}{We replace the $L_1$ loss for super-resolution with a flow-based generalization. The flexibility of our flow-based fidelity loss alleviates the inherent conflict with adversarial losses, leading to a more photo-realistic result and better consistency with the input.
  }%
\label{fig:intro}\vspace{4mm}}

\maketitle

\begin{abstract}

Super-resolution is an ill-posed problem, where a ground-truth high-resolution image represents only one possibility in the space of plausible solutions. Yet, the dominant paradigm is to employ pixel-wise losses, such as $L_1$, which drive the prediction towards a blurry average. This leads to fundamentally conflicting objectives when combined with adversarial losses, which degrades the final quality. We address this issue by revisiting the $L_1$ loss and show that it corresponds to a one-layer conditional flow. Inspired by this relation, we explore general flows as a fidelity-based alternative to the $L_1$ objective. We demonstrate that the flexibility of deeper flows leads to better visual quality and consistency when combined with adversarial losses. We conduct extensive user studies for three datasets and scale factors, where our approach is shown to outperform state-of-the-art methods for photo-realistic super-resolution.\\
Code and trained models will be available at \textcolor{blue}{\href{https://www.git.io/AdFlow}{git.io/AdFlow}}

\end{abstract}

\section{Introduction}

Photo-realistic image super-resolution (SR) is the task of upscaling a low-resolution (LR) image by adding natural-looking high-frequency content. Since this information is not contained in the LR image, SR assumes that a prior can be learned to add plausible high-frequency components. In general, however, there are infinitely many possible high-resolution (HR) images mapped to the same LR image. Therefore, this task is highly ill-posed, rendering the learning of powerful deep SR models highly challenging.

To cope with the ill-posed nature of the SR problem, existing state-of-the-art methods employ an ensemble of multiple losses designed for different purposes~\cite{ledig2017photo,wang2018esrgan,zhang2019ranksrgan}.
In particular, these works largely rely on the $L_1$ loss for fidelity and the adversarial loss for perceptual quality.
Theoretically, the $L_1$ objective aims to predict the average overall plausible HR image manifestations under a Laplace model.
That leads to blurry SR predictions, which are generally not perceptually pleasing. 
In contrast, the adversarial objective prefers images with natural characteristics and high-frequency details. 
These two losses are thus fundamentally conflicting in nature~\cite{BlauM18PerceptionDistTradeoff,blau18PIRM}.

\begin{figure}[t]
    \centering
    \includegraphics[width=1\linewidth]{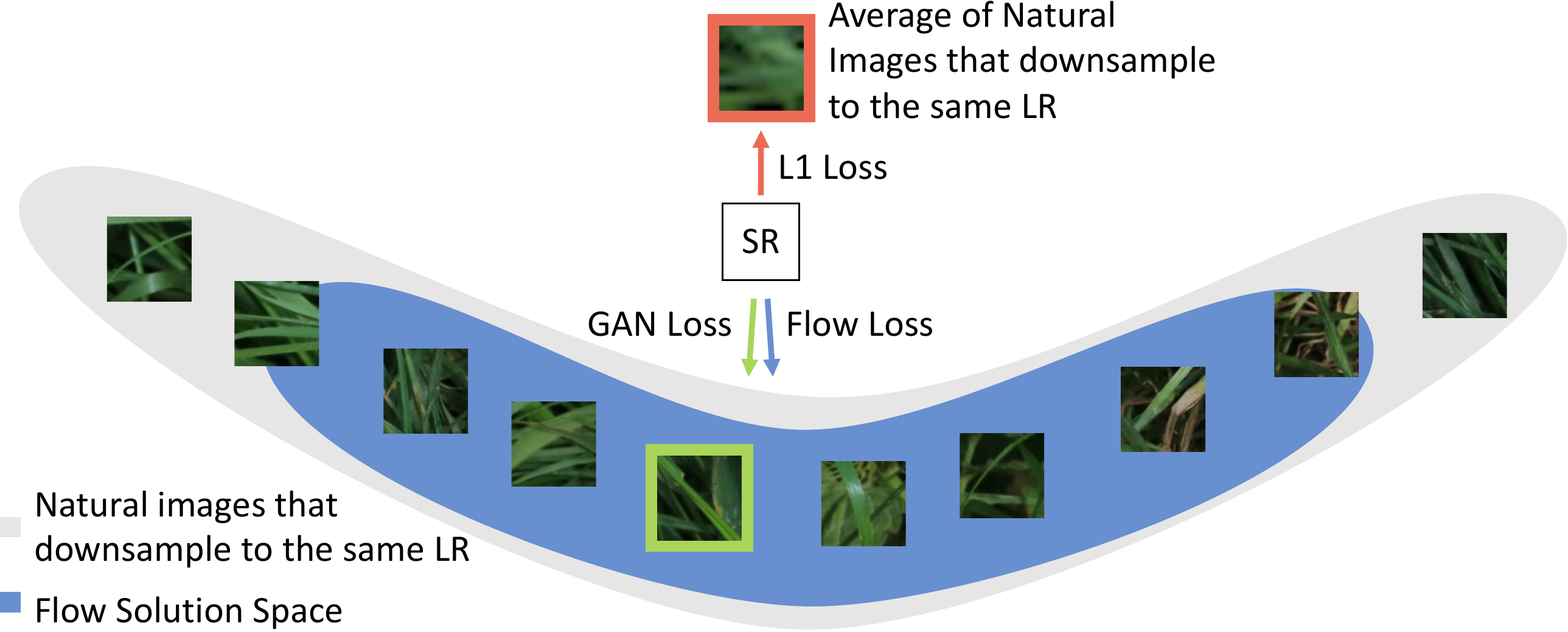}
    \caption{While $L_1$ loss drags the SR prediction towards a blurry mean, both Flow and GAN loss push the prediction towards the real image manifold. Replacing $L_1$ with flow as fidelity term therefore reduces conflict with the GAN loss.}
    \label{fig:influence_of_loss}
    \vspace{-4mm}
\end{figure}

The conflict between the $L_1$ and the adversarial loss has important negative consequences as seen in Figure~\ref{fig:intro}.
In order to find a decent trade-off, a precarious balancing between the two terms is needed.
The found compromise is not optimal in terms of fidelity nor perceptual quality.
Moreover, the conflict between the two losses results in a remarkably inferior low-resolution consistency.
That is, the down-sampled version of the predicted SR image is substantially different from the original LR image.
The conflict between the losses drives the prediction towards a point \emph{outside} the space of plausible HR images (Illustrated in Fig.~\ref{fig:influence_of_loss}).

We attribute those shortcomings to the $L_1$ loss.
Since SR is a highly ill-posed problem, the $L_1$ loss imposes a rigid and exceptionally inaccurate model of the complicated image manifold of solutions. Ideally, we want a loss that ensures fidelity while not penalizing realistic image patches preferred by the adversarial loss. In this work, we therefore first revisit the $L_1$ loss and view it from a probabilistic perspective. We observe that the $L_1$ objective corresponds to a one-layer conditional normalizing flow. That inspires us to explore flow-based generalizations capable of better capturing the manifold of plausible HR images to mitigate the conflict between adversarial and fidelity-based objectives.

A few very recent works \cite{Winkler2020CondNormalizingFlowSR,srflow} have investigated flows for SR.
However, these approaches use heavy-weight flow networks as an \emph{alternative} to the adversarial loss for perceptual quality.
In this work, we pursue a very different view, namely the flow as a fidelity-based generalization of the $L_1$ objective.
Our goal is not to replace the adversarial loss but to find a fidelity-based companion that can enhance the effectiveness of adversarial learning for SR.
In contrast, to~\cite{srflow}, this allows us to employ much shallower and more practical flow networks, ensuring substantially faster training and inference times. Furthermore, we demonstrate that the adversarial loss effectively removes artifacts generated by purely flow-based methods.

\parsection{Contributions}
Our main contributions of this work are as follows:
\textbf{(i)} We revisit the $L_1$ loss from a probabilistic perspective, expressing it as a one-layer conditional flow.
\textbf{(ii)} We generalize the $L_1$ fidelity loss by employing a deep flow and demonstrate that it can be more effectively combined with an adversarial loss.
\textbf{(iii)} We design a more practical, efficient, and stable flow architecture, better suited to the combined objective, leading to $2.5\times$ faster training and inference compared to~\cite{srflow}.
\textbf{(iv)} We perform comprehensive experiments analyzing the flow loss combined with adversarial losses, giving valuable insights on the effects of increasing the flexibility of the fidelity-based objective. In comprehensive user studies, totaling over 50\,000 votes, our approach outperforms state-of-the-art on three different datasets and scale factors.

\section{Related Work}

\parsection{Single Image Super-Resolution} is the task of estimating a  high-resolution image from a low-resolution counterpart. It is fundamentally an ill-posed inverse problem.
While originally addressed by employing interpolation techniques, learned methods are better suited for this complex task.
Early learned approaches used sparse-coding~\cite{DaiTG15JointlyOptimizedRegressorsForSR,SunH12SRFromInternetScaleSceneMatching,YangWHM08SRAsSparseRepresentationOfRawPatches,YangWHM10SRViaSparseRep} and local linear regression~\cite{Timofte2014a+,Timofte13AnchNeighReg,YangY13SimpleFuncSR}. In recent years, deep learning based methods have largely replaced previous techniques for Super-Resolution owing to their highly impressive performance.

Initial deep learning approaches~\cite{DongLHT14LearningDeepConv,dong2016image,kim2016accurate,lai2017deep,lim2017EDSR} for SR aimed at minimize the $L_2$ or $L_1$ distance between the SR and Ground-Truth image. With this objective, the model is effectively trained to predict a mean of plausible super-resolutions corresponding to the given input LR image.
To alleviate this problem~\cite{ledig2017photo} introduced an adversarial and perceptual loss.
Since then, this strategy has remained the predominant approach to super-resolution~\cite{ahn2018image,haris2018deep,Sajjadi17EnhanceNet,wang2018esrgan,lugmayrICCVW2019,fritsche19FrequencySeparation,Ji2020Impressionism,Kim19ProgressFSR}.
Only very few works have investigated other learning formulations. 
Notably, Zhang~\etal~\cite{zhang2019ranksrgan} introduces a selection mechanism based on perceptual quality metrics.
In order to achieve an explorable SR formulation, Bahat~\etal~\cite{bahat2019explorableSR} recently trained a stochastic SR network based on mainly adversarial objectives. The output acts as a prior for a low-resolution consistency enforcing module, optimizing the image in a post-processing step.

In recent works, invertible networks have gained popularity for image-to-image translation \cite{Xiao20InvRescaling,Dinh2017RealNVP,DinhKB14NICENonLinIndependentComponetsEst,RezendeM15VarInferenceWithNF,Winkler2020CondNormalizingFlowSR,srflow,wolf2021deflow,liang2021hierarchical}.
Xiao~\etal~\cite{Xiao20InvRescaling} uses invertible networks to learn down and upscaling of images.
This is similar to compression, but where the compressed representation is constrained to be an LR image.
For super-resolution, \cite{srflow} recently introduced a new strategy based on Normalizing Flows.
It aims at replacing adversarial losses with normalizing flows \cite{Dinh2017RealNVP,DinhKB14NICENonLinIndependentComponetsEst,RezendeM15VarInferenceWithNF}.
In contrast, we investigate conditional flows as a replacement for the $L_1$ loss. In fact, we demonstrate that it forms a direct generalization of the $L_1$ objective.
The aim of this work is to investigate flows as an alternative fidelity-based companion to the adversarial loss.

\section{Method}

\subsection{Revisiting the $L_1$ Loss}
\label{sec:l1}

The standard paradigm for learning a SR network $g$ is to directly penalize the \emph{reconstruction error} between a predicted image $g(x)$ and the ground truth HR image $y \in \reals^{H\times W\times C}$ corresponding to the LR $x$. The reconstruction error is usually measured by applying simple norms in a color space (\eg., RGB or YCbCr). While initial methods \cite{dong2016image,kim2016accurate,lai2017deep} employed the $L_2$ norm, \ie the mean squared error, later works~\cite{ZhaoGFK17L2VsL1, lim2017EDSR} studied the benefit of the $L_1$ error,
\begin{equation}
\label{eq:l1}
    L_1(y, g(x)) = \|y - g(x)\|_1.
\end{equation}
To understand the implications of this objective function, we use its probabilistic interpretation. Namely, that the $L_1$ loss \eqref{eq:l1} corresponds to the Negative Log-Likelihood (NLL) of the Laplace distribution.
This derivation will be particularly illustrative for the generalizations considered later.

We first consider a latent variable $z \in \reals^{H\times W\times C}$ with the standard Laplace distribution $z \sim \mathcal{L}(0,1)$. Let $f$ be a be a function that encodes the LR-HR pair into the latent space as $z = f(y; x) = y - g(x)$. Through the inverse relation $y = f^{-1}(z; x) = z + g(x)$ it is easy to see that $y$ follows a Laplace distribution with mean $g(x)$,
\begin{equation}
    \label{eq:postulLaplace}
    p(y|x; \theta) = \mathcal{L}(y; g(x), 1) = \frac{1}{2^{D}} e^{-\|y - g(x)\|_1} \,.
\end{equation}
Here, $D = HWC$ is the total dimensionality of $y$.
From a probabilistic perspective, we are thus predicting the conditional distribution $p(y|x; \theta)$ of the HR output image $y$ given the LR $x$. In particular, our SR network $g$ estimates the \emph{mean} of this distribution under a Laplacian model. In order to learn the parameters $\theta$ of the network, we simply minimize the NLL $- \log p(y|x; \theta)$ of \eqref{eq:postulLaplace}, which is equal to the $L_1$ loss \eqref{eq:l1} up to an additive constant.

In the aforementioned Laplacian model \eqref{eq:postulLaplace}, derived from the $L_1$ loss \eqref{eq:l1}, only the mean $g(x)$ is estimated from the LR image. Thus, the model assumes that the variance, which reflects the possible variability of each pixel, remains constant. This assumption is however, not accurate. Indeed, super-resolving a constant blue sky is substantially \emph{easier} than estimating the pixel values of a highly textured region, such as the foliage of a tree. In the former case, the predicted pixels should have low variance, while the latter has high variability, corresponding to different possible textures of foliage. For a Laplace distribution, we can encode the variability in the scale parameter $b$, which is proportional to the standard deviation. By predicting the scale parameter $b(x) \in \reals^{H \times W \times C}$ for each pixel, we can learn a more accurate distribution that also quantifies some aspect of the ill-posed nature of the SR problem.

We easily extend our model with the scale parameter prediction by modifying our function $f$ as,
\begin{equation}
\label{eq:scaleL1-f}
    z = f(y;x) = \frac{y - g(x)}{b(x)} \,.
\end{equation}
Since this yields a Laplace distribution $p(y|x;\theta) = \mathcal{L}(y; g(x), b(x))$, we achieve the NLL,
\begin{equation}
\label{eq:scaleL1}
-\log p(y|x; \theta) \propto \left\|\frac{y - g(x)}{b(x)}\right\|_1 +  \sum_{ijc} \log b(x)_{ijc} \,.
\end{equation}
In practice, we can easily modify an SR network to jointly estimating the mean $g(x)$ and scale $b(x)$ by doubling the number of output dimensions. The loss \eqref{eq:scaleL1} stimulates the network to predict larger scale values $b(x)$ for `uncertain' pixels, that are likely to have large error $\|y - g(x)\|$. In principle, \eqref{eq:scaleL1} thus extends the $L_1$ objective to better cope with the ill-posed nature of the SR problem by predicting a more flexible distribution of the HR image. In the next section, we will further generalize the objectives \eqref{eq:l1}, \eqref{eq:scaleL1} through normalizing flows, to achieve an even more flexible fidelity loss.

\subsection{Generalizing the $L_1$ Loss With Flows}
\label{sec:flow}

To capture the probability distribution of the error between the prediction $g(x)$ and ground-truth $y$, we also need to consider spatial dependencies. Neighboring pixels are generally highly correlated in natural images. Indeed, to create coherent textures, even long-range correlations need to be considered. 
However, the $L_1$ loss~\eqref{eq:l1} and its extension \eqref{eq:scaleL1} assume each pixel in $y$ to be conditionally independent given $x$.
In fact, sampling from the predicted conditional distribution $p(y|x;\theta)$ is equivalent to simply adding Laplacian white noise to the predicted mean $g(x)$.
In super-resolution, we strive to create fine textures and details.
To achieve this, the predictive distribution $p(y|x;\theta)$ must capture complex correlations in the image space.

In this paper, we generalize the $L_1$ loss \eqref{eq:scaleL1} with the aim of achieving a more flexible objective, better capturing the ill-posed setting.
That is done through the probabilistic interpretation discussed in Sec.~\ref{sec:l1}.
We observe that the function $f$ introduced in Sec.~\ref{sec:l1} corresponds to a one-layer conditional normalizing flow with a Laplacian latent space.
We can thus generalize this setting by constructing deeper flow networks $f$.
While prior works \cite{Ardizzone19condNFColor,Winkler2020CondNormalizingFlowSR,srflow,pumarola2020cflow} investigate conditional flows for SR as a replacement for adversarial losses, we see it as a generalization of the fidelity-based $L_1$ loss.
With this view, we aim to find a fidelity-based objective better suited for ill-posed problems and, therefore, more effectively combined with adversarial losses.

\begin{figure*}[t]
    \centering
    \includegraphics*[width=1\linewidth,trim=0 25 0 2]{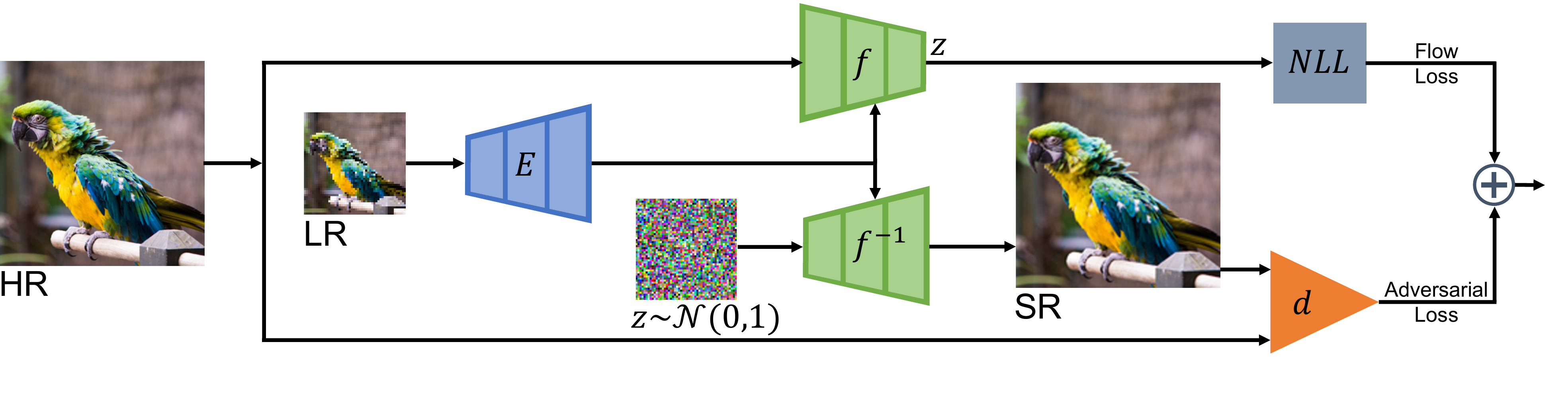}\vspace{-1mm}
    \caption{%
    Overview of our super-resolution approach. 
    Our flow-based NLL loss replaces the often used $L_1$ loss. We accomplish this by encoding the LR image with the network $E(x)$. This conditions the flow $f$, which encodes the GT image $y$. From that, we obtain the NLL loss that drives the SR fidelity. We combine this with a standard adversarial loss, calculated using a discriminator $d$.}\vspace{-3mm}%
    \label{fig:method_architecture}
\end{figure*}

The purpose of the function $f$ is to map the HR-LR pair $(y,x)$ to a latent space $z \sim p_z$, which follows a simple distribution.
By increasing the depth and complexity of the flow $f$, more flexible conditional densities, and therefore also NLL-based losses, are achieved. In the general case, we let flow $f$ to be conditioned on the embedding $E(x)$ of the LR image $x$ as $f(y; x) = f(y; E(x))$. In fact, the network $E$ can be seen as predicting the \emph{parameters} of the conditional distribution $p(y|x;\theta) = p(y|E(x);\theta)$. In this view, the embedding $E(x)$ generalizes the purpose of the SR network $g(x)$, which predicts the mean of the Laplace distribution in the $L_1$ case \eqref{eq:l1}. In the general Laplace case \eqref{eq:scaleL1-f}, the LR embedding network needs to generate both the mean and the scale $E(x) = (g(x), b(x))$. 
Thanks to the flexibility of conditional flow layers, we can however, still use the underlying image representation of any standard SR architecture as $E$. For example, we generate the embedding $E$ by concatenating a series of intermediate feature maps from, \eg, the RRDB~\cite{wang2018esrgan}, or the RCAN~\cite{zhang18RCAN} architecture. For simplicity, we often drop the explicit dependence on $E$ in the flow $f$ and simply write $f(y;x)$.

In order for $f(y;x)$ to be a valid conditional flow network, we need to preserve invertibility in the first coordinate. Under this condition, the conditional density is derived using the change of variable formula~\cite{Dinh2017RealNVP,KingmaD18Glow,srflow} as,
\begin{equation}
    \label{eq:conddens}
    \pyGivenx(\by | \bx, \bT) = \pz\big(\fT(\by; \bx)\big) \left| \det \frac{\partial\fT}{\partial\by}(\by; \bx) \right| \,.
\end{equation}
The latent space prior $z \sim \pz$ is set to a simple distribution, \eg standard Gaussian or Laplacian. The second factor in \eqref{eq:conddens} is the resulting volume scaling, given by the determinant of the Jacobian $\frac{\partial\fT}{\partial\by}$. We can easily draw samples from the model by inverting the flow as $y = f^{-1}(z;x), \, z \sim p_z$. The network $f^{-1}$ thus transforms a simple distribution $p_z$ to capture the complex correlations in the output image space.

The NLL training objective is obtained by applying the negative logarithm to \eqref{eq:conddens},
\begin{subequations}
    \label{eq:nll}
\begin{align}
    \label{eq:nll1}
    - & \!\log \pyGivenx(\by | \bx, \bT) \!=\! 
    -\log \pz(z) \!- \!\log \left| \det \frac{\partial\fT}{\partial\by}(\by; \bx) \right| \!\! \\ \label{eq:nll2}
    &= -\log \pz(\bz) - \sum_{n=0}^{N-1} \log \left| \det \frac{\partial\fT^n}{\partial\bh^{n}}(\bh^{n}; E(\bx)) \right| \,,
\end{align}
\end{subequations}
where $z = \fT(\by; \bx)$.
In the second equality, we have decomposed $\fT$ into the sequence of $N$ flow layers $h^{n+1} = \fT^n(h^n; E(x))$, with $h^0 = y$ and $h^{N} = z$.
This allows for efficient computation of the log-determinant term.

We can now derive that the flow objective \eqref{eq:nll} generalizes the scaled $L_1$ loss \eqref{eq:scaleL1} and thereby also the standard $L_1$ loss \eqref{eq:l1}. By using the function $f$ defined in \eqref{eq:scaleL1-f} and the standard Laplacian latent variable $\pz(z) = \mathcal{L}(z; 0, 1)$, we derive the first term in \eqref{eq:scaleL1} is by inserting \eqref{eq:scaleL1-f} into the first term $-\log \pz(z)$ in \eqref{eq:nll1}. For the second term, we first immediately obtain the Jacobian of \eqref{eq:scaleL1-f} as a diagonal matrix $\frac{\partial f}{\partial y} = \text{diag}(\frac{1}{b(x)})$ with elements $\frac{1}{b(x)_{ijk}}$. Inserting this result into the log-determinant term in \eqref{eq:nll1} yields the second term in \eqref{eq:scaleL1}. A more detailed derivation is provided in the supplementary material. Next, we employ the flow-based fidelity objective in a full super-resolution framework by combining it with adversarial losses.

\subsection{Flow-Fidelity with Adversarial Losses}
\label{ssec:combinationOfGAN}

The introduction of adversarial losses~\cite{ledig2017photo} pioneered a new direction in super-resolution, aiming to generate perceptually pleasing HR outputs from the natural image manifold.
In order to achieve this, the adversarial loss needs to be combined with fidelity-based objectives, ensuring that the generated SR image is close to the HR ground-truth.
Therefore, SRGAN~\cite{ledig2017photo} and later works \cite{ahn2018image,haris2018deep,Sajjadi17EnhanceNet,lugmayrICCVW2019,fritsche19FrequencySeparation, Ji2020Impressionism} most typically combine the adversarial loss with the $L_1$ objective.
However, these two objectives are fundamentally conflicting.
Unlike the $L_1$ loss that pulls the super-resolution towards the mean of all plausible manifestations, the adversarial loss forces the generator to choose exactly one image of the natural image manifold.
Hence, the adversarial objective ideally assigns a low loss on all natural image patches.
In contrast, such predictions generate a high $L_1$ loss since it prefers the blurry average of plausible predictions.
We aim to resolve this issue by replacing the $L_1$ loss with the aforementioned flow-based generalizations.

The flow can learn a more flexible conditional distribution $p(y|x;\theta)$ of the HR $y$. It therefore better spans the natural image manifold while simultaneously encouraging consistency with the input LR image $x$. The NLL loss \eqref{eq:nll} of the flow distribution does therefore not penalize patches from the natural image manifold to the same extent. That allows the adversarial objective to drive the generated SR images towards perceptually pleasing results without being penalized by the fidelity-based loss. Conversely, the flow-based fidelity loss allows the network to learn from the single provided ground-truth HR image $y$, without reducing perceptual quality or incurring a higher adversarial loss. 

Interestingly, the flow network $y = \fT^{-1}(z;x), \, z \sim p_z$ can also be seen as stochastic generator for the adversarial learning. While stochastic generators are fundamental to unconditional Generative Adversarial Networks (GANs) \cite{Goodfellow14GAN}, deterministic networks are most common in the conditional setting, including super-resolution. In fact, GANs are well known to be highly susceptible to mode collapse in the conditional setting \cite{IsolaZZE17pix2pix,MathieuCL15VideoMSE}. In contrast, flows are highly resistant to mode collapse due to the bijective constraint on $\fT$. This is highly important for ill-posed problems such as SR, where we ideally want to span the space of possible predictions. However, it is important to note that the flow is not merely a generator, as in the standard GAN setting. The flow itself also serves as a flexible loss function \eqref{eq:nll}.

Formally, we add the adversarial loss on samples generated by the flow network $f$. Let $d_\phi$ be the discriminator with parameters $\phi$. For one LR-HR pair $(x,y)$ and random latent sample $z \sim p_z$, we consider the adversarial loss
\begin{equation}
    \label{eq:gan}
    L_\text{adv} = \log\left(1 - d_\phi\big(\fT^{-1}(z;x)\big)\right) + \log\big(d_\phi(y)\big)\,.
\end{equation}
The loss \eqref{eq:gan} is minimized \wrt the flow and LR encoder parameters $\theta$ and maximized \wrt the discriminator parameters $\phi$. In general, any other variant of adversarial loss \eqref{eq:gan} can be employed. During training, we employ a linear combination of the NLL loss \eqref{eq:nll} and \eqref{eq:gan}. Our training procedure is detailed in Sec.~\ref{sec:training}.

\subsection{Conditional Flow Architecture}
\label{ssec:architecture_details}

Our full approach, depicted in Fig.~\ref{fig:method_architecture}, consists of the super-resolution network $E$, the flow network $f$ and the discriminator $d$.
We construct our conditional flow network $f$ based on~\cite{srflow} and use the same settings, where not mentioned otherwise. It is based on Glow~\cite{KingmaD18Glow} and RealNVP~\cite{Dinh2017RealNVP}.
It employs a pyramid structure with $L$ scales, each halving the previous layer's spatial size using a squeeze layer and, depending on the number of channels, also bypassing half of the activations directly to the NLL calculation.
We use 3, 4, and 4 scale levels for $4\times$, $6\times$ and $8\times$ respectively, each consisting of a series of $K$ flow steps. 

Each Flow-Step consists of a sequence of four layers.
In encoding direction, we first employ the ActNorm~\cite{KingmaD18Glow} to normalize the activations using a learned channel-wise scale and bias.
To establish information transfer across the channel dimension, we then use an invertible $1 \times 1$ convolution~\cite{KingmaD18Glow}.
The following layers condition the flow on the LR image similar to~\cite{srflow}.
First, the Conditional Affine Coupling~\cite{Dinh2017RealNVP,srflow},  partitions the channels into two halves.
The first half is used as input, together with the LR encoding $E(x)$, to a 3-layer convolutional network module, which predicts the element-wise scale and bias for the second half.
This module adds non-linearities and spatial dependencies to the flow network while ensuring easy invertibility and tractable log-determinants.
Secondly, the Affine Image Injector is applied, which transforms all channels conditioned on the low-resolution encoding $E(x)$.

Instead of the learnable $1 \times 1$ convolutions used in \cite{srflow,KingmaD18Glow}, we use constant orthonormal matrices that are randomly sampled at start of the training. 
We found this to significantly improve training speed while ensuring better stability due to these layers' perfect conditioning. 
When combined with an adversarial loss, the flow network operates in both the encode $\fT$ and decode $\fT^{-1}$ direction during training. To ensure stability during training in both directions, we reparametrize the prediction of the multiplicative unit in the conditional affine coupling layer. In particular, we predict the multiplicative factor as $s = \text{Sigmoid}(\tilde{s})^{-1}$, where $\tilde{s}$ is the unconstrained prediction stemming from the convolutional module in the coupling.

\newcommand{\saturation}{0.9}
\newcommand{\brightness}{0.7}

\definecolor{losecolor}{hsb}{0.98, \saturation, \brightness}
\definecolor{wincolor}{hsb}{0.3, \saturation, \brightness}
\definecolor{tiecolor}{hsb}{0, 0, \brightness}

\newcommand{\lose}[1]{\textcolor{losecolor}{#1}}
\newcommand{\win}[1]{\textcolor{wincolor}{#1}}
\newcommand{\tie}[1]{\textcolor{tiecolor}{#1}}

\begin{table*}[t]
    \centering%
    \resizebox{\linewidth} {!}{%
    \begin{tabular}{l | ccc | ccc | ccc}
        \toprule
        AdFlow & \multicolumn{3}{c}{\textbf{4$\times$}} & \multicolumn{3}{c}{\textbf{6$\times$}} & \multicolumn{3}{c}{\textbf{8$\times$}} \\
        compared to       & DIV2K         & BSD           & Urban         & DIV2K         & BSD           & Urban         & DIV2K         & BSD           & Urban         \\
        \midrule
        BaseFlow     & \win{62.1\% ± 2.2} & \win{68.3\% ± 2.4} & \win{74.2\% ± 2.1} & \win{73.4\% ± 2.0} & \win{80.7\% ± 1.8} & \win{82.9\% ± 1.7} & \win{69.2\% ± 2.1} & \win{73.1\% ± 2.0} & \win{78.2\% ± 1.9} \\
        SRFlow       & \win{60.1\% ± 2.2} & \win{67.2\% ± 2.4} & \win{66.3\% ± 2.2} & \NA           & \NA           & \NA           & \win{66.2\% ± 2.1} & \win{67.2\% ± 2.1} & \win{71.8\% ± 2.0} \\
        RankSRGAN    & \win{56.9\% ± 2.2} & \win{54.8\% ± 2.6} & \win{67.5\% ± 2.2} & \NA           & \NA           & \NA           & \NA           & \NA           & \NA           \\
        ESRGAN       & \win{56.1\% ± 2.2} & \win{51.2\% ± 2.6} & \win{64.5\% ± 2.3} & \win{57.5\% ± 2.3} & \win{62.8\% ± 2.2} & \win{63.8\% ± 2.2} & \tie{49.9\% ± 2.3} & \win{54.5\% ± 2.2} & \win{57.1\% ± 2.2} \\
        Ground Truth & \tie{49.0\% ± 2.2} & \lose{25.4\% ± 2.2} & \lose{29.1\% ± 2.2} & \lose{27.4\% ± 2.1} & \lose{8.9\% ± 1.3} & \lose{11.6\% ± 1.5} & \lose{18.3\% ± 1.7} & \lose{4.2\% ± 0.9} & \lose{7.7\% ± 1.2} \\
        \bottomrule
    \end{tabular}
    }
    \caption{
        Quantitative results of the user study. Each entry is aggregated from 1\,500 votes. For each dataset and scale factor, we directly compare AdFlow with each competing method in a pairwise fashion, as detailed in Sec.~\ref{sec:sota}.
        For each compared method (left), we report the proportion of votes \emph{in favor} of AdFlow along with the 95\% confidence interval.
        We indicate if AdFlow is significantly \win{better} or \lose{worse}.
    }\vspace{-3mm}%
    \label{tab:mturk}
    \vspace{0mm}
\end{table*}

\parsection{Super-Resolution embedding network $E(x)$} Our flow-based objective is designed as a replacement of $L_1$ loss. Our formulation is therefore agnostic to the architecture underlying SR embedding network $E$.
We use the popular RRDB~\cite{wang2018esrgan} SR network as our encoder $E$.
Instead of outputting the final RGB SR image, these networks predict a rich embedding of the LR image.
We obtain this in practice by simply concatenating the underlying feature activations at the intermediate RRDB blocks 1, 4, 6 and 8.

\parsection{Discriminator}
We use the VGG-based network from~\cite{wang2018esrgan} as a discriminator. Since we generate stochastic SR samples during training, we found it beneficial to reduce the discriminator's capacity to ensure a balanced adversarial objective. We, therefore, reduce the internal channel dimension of the discriminator from 64 to 16.

\subsection{Training Details}
\label{sec:training}
Our approach is trained by a weighted combination of the NLL loss \eqref{eq:nll} for fidelity and the adversarial loss \eqref{eq:gan} to increase perceptual quality. We consider the standard bicubic setting in our experiments, where the LR is generated with the MATLAB bicubic downsampling kernel. In particular, we train for both $4\times$ and the challenging $8\times$ SR scenario. We first train the networks $E$ and $f$ using only the flow NLL loss for 200k iterations using initial learning rates of $10^{-5}$ for $4\times$ and $10^{-6}$ for $8\times$, which are then decreased step-wise. 
We fine-tune the network with the adversarial loss for 200k iterations and select the checkpoint with lowest LPIPS~\cite{zhang2018unreasonable} as measured the training set. 
We employ the Adam~\cite{KingmaB14Adam} optimizer. 
As in~\cite{srflow} we add uniformly distributed noise with a strength of $\frac{1}{32}$ of the signal range to the ground-truth HR.
Our network is trained on HR patches of $160 \times 160$ pixels for $4\times$ and $8\times$ and $144 \times 144$ pixels for $6\times$.
In principle, our framework can employ any adversarial loss formulation.
To allow for a direct comparison with the popular state-of-the-art network ESRGAN~\cite{wang2018esrgan}, we employ the same relativistic adversarial formulation. For $4\times$ and $6\times$ SR, we weight the adversarial loss with a factor of $10^{-2}$ and use a discriminator learning rate of $10^{-3}$.
For $8\times$, we use $0.1$ and $10^{-4}$ respectively
We use the same training data employed by ESRGAN~\cite{wang2018esrgan}, consisting of the DF2K dataset. It comprises 2650 training images from Flickr2K~\cite{lim2017EDSR} and 800 training images from the DIV2K~\cite{div2k} dataset.

\section{Experiments}

\begin{figure*}[t]
\rotatebox[origin=c]{90}{\resizebox{5.7cm}{!}{
		\begin{tabular}{ C{1.5cm} C{1.5cm} C{3cm}}
			Urban100 & BSD100 & DIV2K
		\end{tabular}
}}\vspace{0mm}%
\begin{minipage}{0.95\linewidth}%
\includegraphics[width=\linewidth]{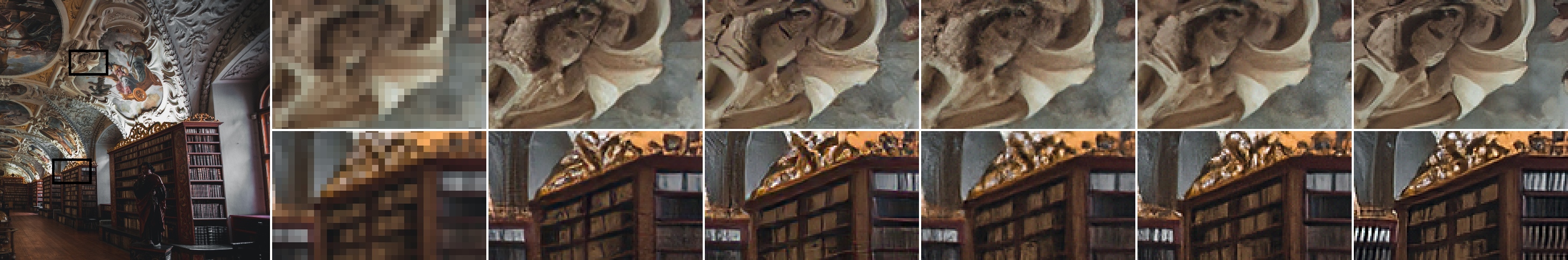}
\includegraphics[width=\linewidth]{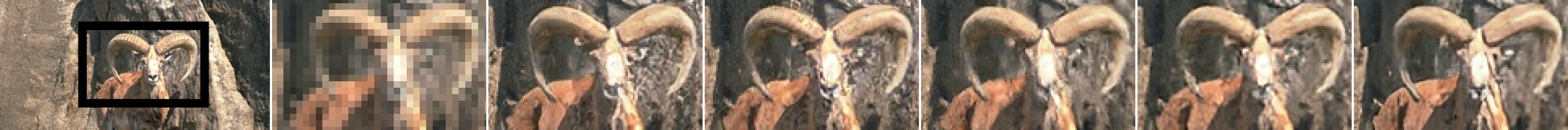}
\includegraphics[clip, trim=0 0 0 243, width=\linewidth]{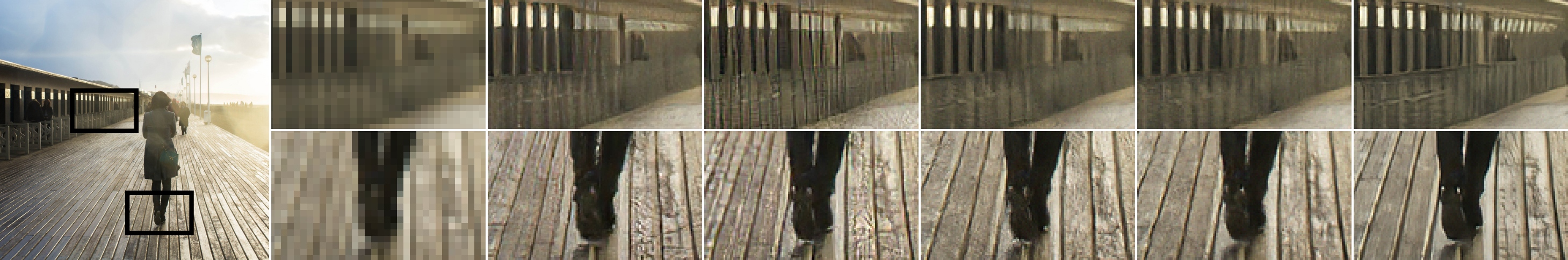}
\end{minipage}
\resizebox{1.00\linewidth}{!}{%
\begin{tabular}{ C{4.2cm} C{3cm} C{3cm} C{3cm} C{3cm} C{3cm} C{3cm} C{0.5cm} }
& Low Resolution & RankSRGAN & ESRGAN\cite{wang2018esrgan} & SRFlow\cite{srflow} & BaseFlow & AdFlow &
\end{tabular}}\vspace{-1mm}
\caption{Qualitative comparison with state-of-the-art approaches on the DIV2K (val), BSD100 and Urban100 set for $4\times$ SR.}
\vspace{-2mm}
\label{fig:sota_x4}
\end{figure*}

\begin{figure*}[t]
\rotatebox[origin=c]{90}{\resizebox{8cm}{!}{
		\begin{tabular}{ C{2cm} C{2cm} C{4cm}}
			Urban100 & BSD100 & DIV2K
		\end{tabular}
}}\vspace{0mm}%
\begin{minipage}{0.95\linewidth}
\includegraphics[width=\linewidth]{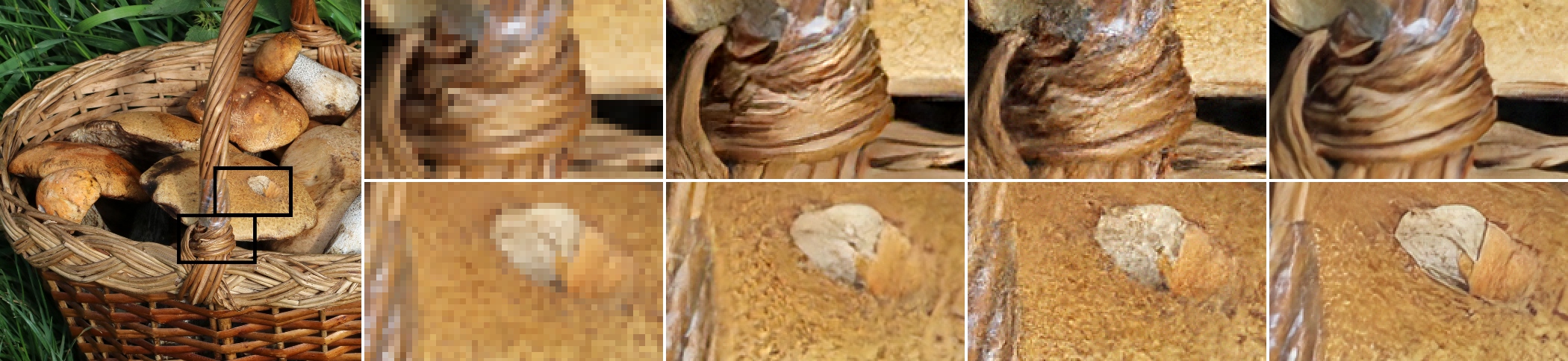}
\includegraphics[width=\linewidth]{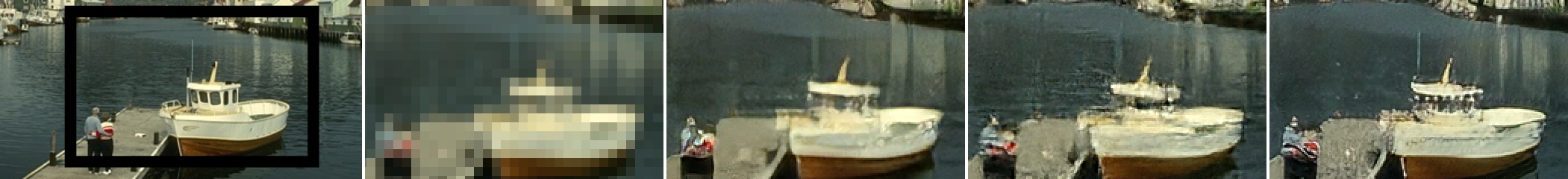}
\includegraphics[width=\linewidth]{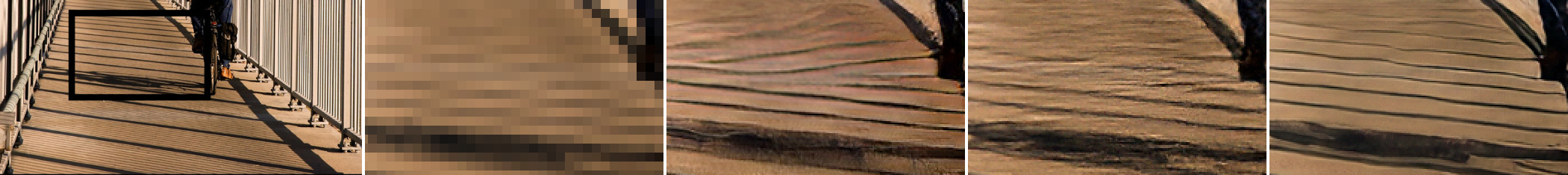}
\end{minipage}
\resizebox{1.00\linewidth}{!}{%
\begin{tabular}{ C{4.2cm} C{3cm} C{3cm} C{3cm} C{3cm} C{0.15cm} }
& Low Resolution & ESRGAN\cite{wang2018esrgan} & BaseFlow & AdFlow &
\end{tabular}}\vspace{-1mm}
\caption{Qualitative comparison with state-of-the-art approaches on the DIV2K (val), BSD100 and Urban100 set for $6\times$ SR.}
\label{fig:sota_x6}
\vspace{-2.5mm}
\end{figure*}

\begin{figure*}[t]
\rotatebox[origin=c]{90}{\resizebox{6.5cm}{!}{
		\begin{tabular}{ C{1.5cm} C{1.5cm} C{3cm}}
			Urban100 & BSD100 & DIV2K
		\end{tabular}
}}\vspace{0mm}%
\begin{minipage}{0.95\linewidth}
\includegraphics[width=\linewidth]{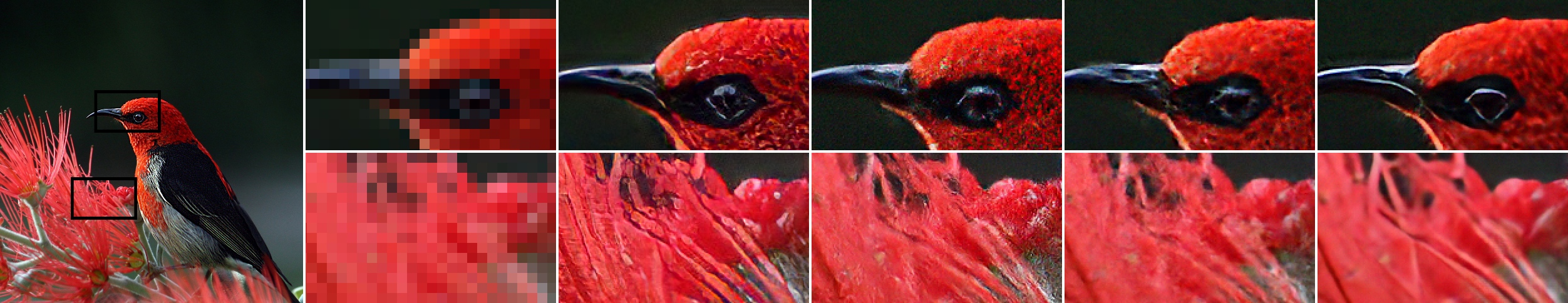}
\includegraphics[width=\linewidth]{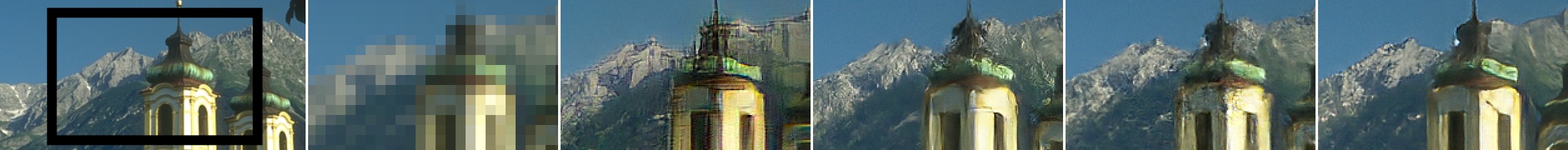}
\includegraphics[clip, trim=0 243 0 0, width=\linewidth]{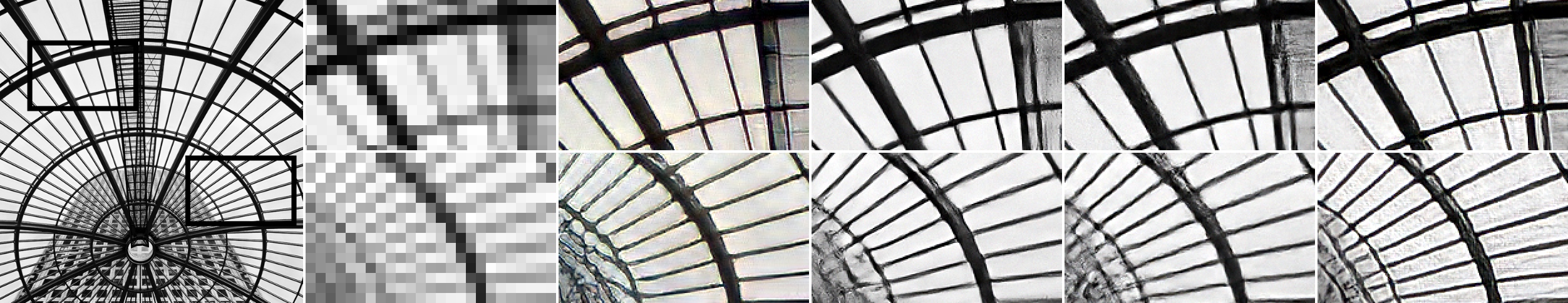}
\end{minipage}
\resizebox{1.00\linewidth}{!}{%
\begin{tabular}{ C{3.9cm} C{3cm} C{3cm} C{3cm} C{3cm} C{3cm} C{0.2cm} }
& Low Resolution & ESRGAN\cite{wang2018esrgan} & SRFlow\cite{srflow} & BaseFlow & AdFlow &
\end{tabular}}\vspace{-1mm}
    \caption{Qualitative comparison with state-of-the-art approaches on the DIV2K (val), BSD100 and Urban100 set for $8\times$ SR.}
\label{fig:sota_x8}
\vspace{-5mm}
\end{figure*}

We validate our proposed formulation by performing comprehensive experiments on the three standard datasets,
namely DIV2K~\cite{div2k}, BSD100~\cite{BSD100} and Urban100~\cite{Urban100}.
We train our approach for three different scale factors $4\times$, $6\times$, and $8\times$.
We term our flow-only baseline as \textbf{BaseFlow} and our final method, which also employs adversarial learning, as \textbf{AdFlow}.
The prediction and evaluation is performed on the full image resolution.
For the purely flow-based baseline, we found it best to use a sampling temperature~\cite{KingmaD18Glow,srflow} of $0.9$.
For our final approach with adversarial loss, we found the standard sampling temperature of $1.0$ to yield best results.
Detailed results and more visual examples are found in the supplementary material.

\subsection{State-of-the-Art Comparison}
\label{sec:sota}

We first compare our approach with state-of-the-art.
This work aims to achieve SR predictions that are (i) photo-realistic and (ii) consistent with the input LR image.
Since it has become well known~\cite{ledig2017photo,Sajjadi17EnhanceNet,wang2018esrgan,lugmayrICCVW2019,fritsche19FrequencySeparation,Ji2020Impressionism,srflow,AIM2019RWSRchallenge,lugmayr2021ntire,NTIRE2020RWSRchallenge} that computed metrics, such as PSNR and SSIM, fail to rank methods according to photo-realism (i), we therefore perform extensive user studies as further described below.
To assess the consistency of the prediction with the LR input (ii), we first downscale the predicted SR image with the given bicubic kernel and compare the result with the LR input.
Their similarity is measured using PSNR, and we therefore refer to this metric as LR-PSNR.
The LR consistency penalizes hallucinations and artifacts that cannot be explained from the input image.

\parsection{User studies}
We compare the photo-realism of our AdFlow with other methods in user studies. The user is shown the full low-resolution image where a randomly selected region is marked with a bounding box.
Next to this image, two different super-resolutions, or ``zooms'', of the marked region, are displayed.
The user is asked to select ``Which image zoom looks more realistic?''.
In this manner, the user evaluates the photo-realism of our AdFlow versus each compared method.
To obtain an unbiased opinion, the methods were anonymized to the user and shown in a different random order for each crop.
In each study, we evaluate 3 random crops for each of the 100 images in a dataset (DIV2k, BSD100, or Urban100).
We use 5 different users for every study, resulting in 1500 votes per method-to-method comparison in each dataset and scale factor.
The full user study is shown in Tab.~\ref{tab:mturk}, thus collects over 50\,000 votes.
Further details are provided in the supplement.

\parsection{Methods}
We compare our approach with state-of-the-art approaches for photo-realistic super-resolution: ESRGAN~\cite{wang2018esrgan}, RankSRGAN~\cite{zhang2019ranksrgan}, and SRFlow~\cite{srflow}.
For the two latter approaches, we use the publicly available code and trained models ($4\times$ for RankSRGAN, $4\times$ and $8\times$ for SRFlow).
In addition to the publicly ESRGAN model for $4\times$ SR, we train models for $6\times$ and $8\times$ SR using the code provided by the authors.
All compared methods are trained on the same training set, namely DF2k~\cite{lim2017EDSR}.

\parsection{Results}
The results of our user study are given in Table.~\ref{tab:mturk}.
The first number represents the ratio of votes in favour of AdFlow and the second the 95\% confidence interval.
AdFlow outperforms all other methods with significance level 95\% in all datasets except for one case.
Interestingly, it almost matches the realism of the ground-truth on the DIV2k dataset.
The visual results in Fig.~\ref{fig:sota_x4} show that our AdFlow generates sharp and realistic textures and structures.
In contrast, ESRGAN frequently generates visible artifacts while SRFlow and BaseFlow achieve less sharp results.
While RankSRGAN experiences fewer artifacts compared to ESRGAN, its predictions are less sharp compared to AdFlow.

\newcommand{\STAB}[1]{\begin{tabular}{@{}c@{}}#1\end{tabular}}

\begin{table}[b]
    \centering\vspace{-5mm}%
    \resizebox{\linewidth} {!}{%
    \begin{tabular}{l l | c@{~~}c@{~~}c | c@{~~}c@{~~}c | c@{~~}c@{~~}c}
        \toprule
        & & \multicolumn{3}{c}{$4\times$} & \multicolumn{3}{c}{$6\times$} & \multicolumn{3}{c}{$8\times$} \\
        & Method    & DIV2K & BSD  & Urban & DIV2K & BSD  & Urban & DIV2K & BSD  & Urban \\
        \midrule
        \multirow{2}{*}{\STAB{\rotatebox[origin=c]{90}{\STAB{pure \\ Flow}}}}
        & BaseFlow           & 49.9            & \textbf{49.9} & \textbf{49.5}  & \textbf{48.3}  & \textbf{48.6} & \textbf{47.9}  & \textbf{49.8}            & 50.2 & \textbf{48.7}  \\
        & SRFlow             & \textbf{50.0}   & \textbf{49.9} & \textbf{49.5}  & \NA   & \NA  & \NA   & 49.0            & \textbf{51.0} & 48.1  \\
        \midrule
        \multirow{3}{*}{\STAB{\rotatebox[origin=c]{90}{\STAB{Adv. \\ combo}}}}
        & RankSRGAN          & 42.3            & 41.7 & 39.9  & \NA   & \NA  & \NA   & \NA             & \NA  & \NA   \\
        & ESRGAN    & 39.0  & 37.7 & 36.8  & 33.2  & 32.8 & 30.9  & 31.3  & 31.7 & 28.9  \\
        & \textbf{AdFlow}    & \textbf{45.2}  & \textbf{45.6} & \textbf{43.4}  & \textbf{37.5}  & \textbf{38.5} & \textbf{36.0}  & \textbf{46.0}  & \textbf{46.8} & \textbf{42.1}  \\
        \bottomrule
    \end{tabular}
    }
    \caption{
        Consistency to the input in terms of LR-PSNR (dB) on the DIV2K (val), BSD100 and Urban100 datasets.
        We compare for methods that employ adversarial loss (bottom).
        While purely Flow based methods (top) achieve high LR-PSNR, they have worse perceptual quality (Tab.~\ref{tab:mturk}).
        AdFlow outperforms other methods using adversarial loss for LR consistency significantly.}%
    \label{tab:lrpsnr}
\end{table}

The results for higher scale factors $6\times$ and $8\times$ show a similar trend as seen in Fig.~\ref{fig:sota_x6} and~\ref{fig:sota_x8}. 
Our approach consistently outperforms the purely flow-based approaches BaseFlow and SRFlow for all scale factors and datasets by over $20\%$ of the votes.
As seen in the visual examples, particularly for $6\times$ and $8\times$, the flow-based approaches often generate strong high-frequency artifacts. In contrast, our AdFlow generates structured and crips textures attributed to the adversarial loss.
Compared to ESRGAN, which combines $L_1$ with adversarial loss, AdFlow produces generally sharper results and has no visible color shift, as seen in Fig.~\ref{fig:sota_x8}.
Interestingly, AdFlow demonstrates substantially better generalization to the BSD100 and Urban100 datasets than ESRGAN, as shown in the user study in Table~\ref{tab:mturk}.
This indicates that ESRGAN tends to overfit to the DIV2k distribution. 
Qualitative examples for $4\times$, $6\times$, and $8\times$ are shown in Fig.~\ref{fig:sota_x4}, \ref{fig:sota_x6}, and \ref{fig:sota_x8}, respectively.

We report the LR-PSNR for all datasets and scale factors in Tab.~\ref{tab:lrpsnr}.
ESRGAN and RankSRGAN obtain poor LR consistency across all datasets, as shown in Tab.~\ref{tab:lrpsnr}.
AdFlow gains $4.3$dB - $15.1$dB in LR-PSNR over ESRGAN, indicating less hallucination artifacts and color shift.
By employing flow-based fidelity instead of $L_1$, AdFlow achieves superior photo-realism while ensuring high LR consistency.

\subsection{Analysis of Flow-based Fidelity Objective}
\begin{figure}[t]
    \centering%
    \includegraphics*[width=\linewidth,trim=0 0 0 8]{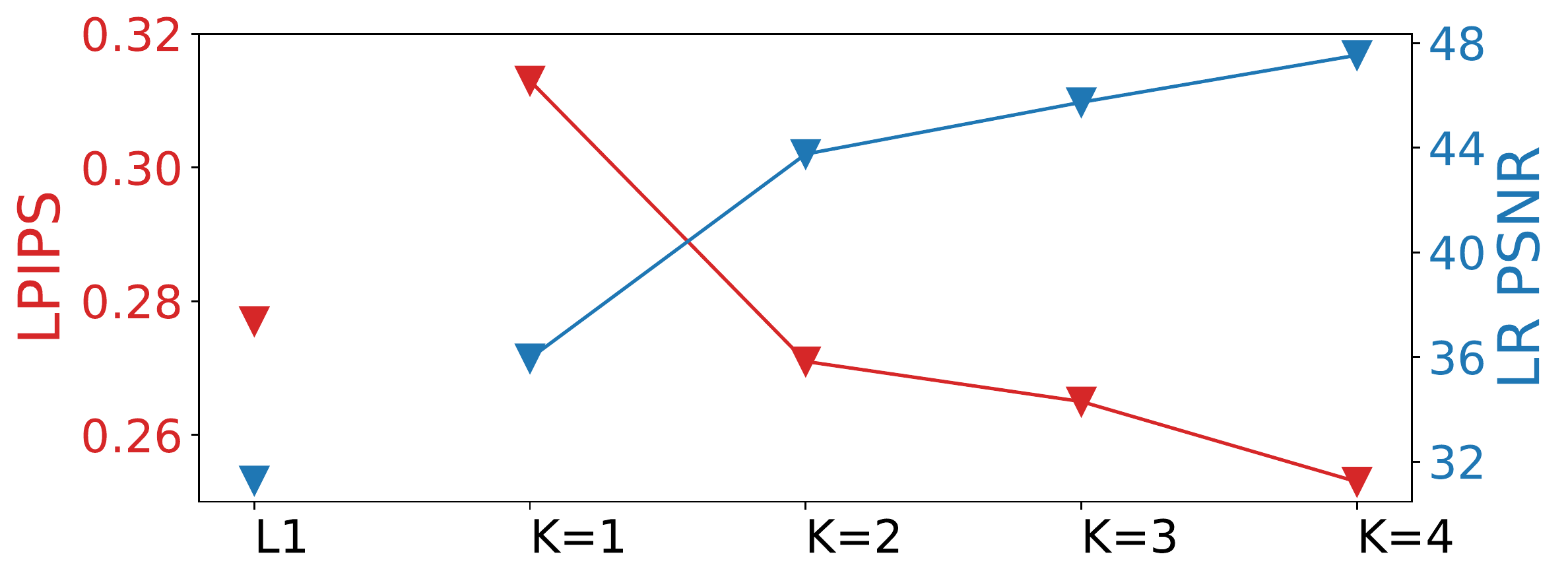}\vspace{-2mm}
    \caption{Analysis of the input consistency LR-PSNR and perceptual quality LPIPS for different numbers of Flow Steps $K$.}\vspace{-1.5mm}
    \label{fig:ablation_k_plot}
    \vspace{-3.5mm}
\end{figure}

Here, we analyze the impact of generalizing the $L_1$ loss towards a gradually more flexible flow-based NLL objective.
This is done by increasing the number of flow steps $K$ per level inside the flow architecture.
We train and evaluate our AdFlow with different depths $K$ for $8\times$ SR. Due to the difficulty and cost of running a large number of user studies, we here use the learned LPIPS~\cite{zhang2018unreasonable} distance as a surrogate to assess photo-realism. %
In Fig.~\ref{fig:ablation_k_plot} we plot the LPIPS and LR-PSNR on the DIV2K validation set \wrt the number of flow-steps $K$. We also include the results obtained by the $L_1$ loss, which is an even simpler one-layer flow loss, as discussed in Sec.~\ref{sec:l1} and \ref{sec:flow}. Note that these results correspond to the standard RRDB and ESRGAN, respectively.

\begin{table}[b]
    \centering\vspace{-3mm}%
    \resizebox{\linewidth}{!}{%
    
\vspace{-4mm}
\begin{tabular}{@{~}c@{~~~}c@{~~~}c@{~~~}c@{~~~}c@{~~~}c@{~~~}c@{~}}
\toprule
Adv. Loss & Affine Coup. &  Rand. Rot. & Coupl. Mult.  & Percept. Loss  & LPIPS~$\downarrow$ & LR-PSNR~$\uparrow$ \\
\midrule                                                                             
           &               &                  &               &             &  0.349 &   39.76 \\
\checkmark &               &                  &               &             &  0.337 &   34.85 \\
           & \checkmark    &                  &               &             &  0.253 &   50.16 \\
\checkmark & \checkmark    &                  &               &             &      - &       - \\
           & \checkmark    & \checkmark       &               &             &  0.254 &   50.19 \\
\checkmark & \checkmark    & \checkmark       &               &             &      - &       - \\
           & \checkmark    & \checkmark       & \checkmark    &             &  0.253 &   49.78 \\
\checkmark & \checkmark    & \checkmark       & \checkmark    &             &  0.253 &   47.54 \\
\checkmark & \checkmark    & \checkmark       & \checkmark    & \checkmark  &   0.270 &              47.35 \\
\bottomrule
\end{tabular}

    }
    \caption{Ablation of architecture choice for adversarial loss (Adv.), the use of Affine Couplings (Aff. Coup.), Random Depth-Wise Rotation (Rand. Rot.), Decode Multiplication for the Affine Couplings (Decode Mult.) and perceptual loss (Percept.)~\cite{ledig2017photo}.}%
    \label{tab:ablation_permRot_DecodeMult_div2k_8x}%
\end{table}

As we increase the depth $K$ of the flow network $f$, the LPIPS decreases while the LR-PSNR increases. This indicates an improvement in perceptual quality and low-resolution consistency.
This trend also holds when starting from the $L_1$ NLL objective. Note that the brief increase in LPIPS is explained by the added stochasticity when transitioning from the $L_1$ to the $K=1$ flow. Indeed, a too shallow flow network does not capture rich enough spatial correlations in order to generate more natural samples. However, already at $K=2$, the flow-based generalization outperforms the $L_1$ in LPIPS. Increasing the flexibility of the NLL-based fidelity loss, starting from $L_1$, thus benefits perceptual quality and consistency. This strongly indicates that a flow-based fidelity objective alleviates the conflicts between the adversarial loss and $L_1$ loss.

\subsection{Ablation of Flow Architecture}

In Tab.~\ref{tab:ablation_permRot_DecodeMult_div2k_8x} we show results of our ablative experiments for $8\times$ SR on DIV2k.
First, we ablate the use of Conditional Affine Couplings. Removing this layer (top row) results in a conditionally linear flow, thereby radically limiting its expressiveness. This leads to a substantially worse LPIPS and LR-PSNR, demonstrating the importance of a flexible flow network. 
Second, we replace the learnable $1 \times 1$ convolutions with fixed random rotation matrices (Rand.~Rot.). While widely preserving the quality in all metrics, it reduces the training time by $37.1\%$. Next, we consider the reparametrization of the coupling layers (Coupl.~Mult.). We found this to be critical for training stability when combined with the adversarial loss. Lastly, we investigate the use of the VGG-based perceptual loss~\cite{ledig2017photo} that is commonly used in SR methods. It is generally employed as a more perceptually inclined fidelity loss to complement the $L_1$ objective. However, we found the perceptual loss not to be beneficial. This indicates that the more flexible flow-based fidelity loss can also effectively replace the VGG loss.

\section{Conclusion}
We explore conditional flows as a generalization of the $L_1$ loss in the context of photo-realistic super-resolution. In particular, we tackle the conflicting objectives between $L_1$ and adversarial losses. Our flow-based alternatives offer both improved fidelity to the input low-resolution and a higher degree of flexibility. Extensive user studies clearly demonstrate the advantages of our approach over state-of-the-art on three datasets and scale factors. Lastly, our experimental analysis brings new insights into the learning of super-resolution methods, paving for further explorations in the pursuit of more powerful learning formulations.

\parsection{Acknowledgements}
This work was supported by the ETH Z\"urich Fund (OK), a Huawei Technologies Oy (Finland) project and an Nvidia GPU grant and AWS.

{\small
\bibliographystyle{ieee_fullname}
\bibliography{references}
}

\clearpage

\setcounter{section}{0}
\renewcommand{\thesection}{\Alph{section}}

\section*{Appendix}

In this appendix, we first provide further details on the user study in Sec.~\ref{sec:userStudy}.
Secondly, we provide an analysis of the sampling temperature in Sec~\ref{sec:percDist}.
Third, we present additional details about the minimally generalized $L_1$ loss in Sec.~\ref{sec:varL1}.
Finally, we provide a further qualitative and quantitative comparison of AdFlow with other state-of-the-art methods in Sec.~\ref{sec:extendedEval}.
Additional visual results, used in our study, will be available on the project page  \textcolor{blue}{\href{https://www.git.io/AdFlow}{git.io/AdFlow}}.

\begin{figure}[b]
    \includegraphics[width=\linewidth]{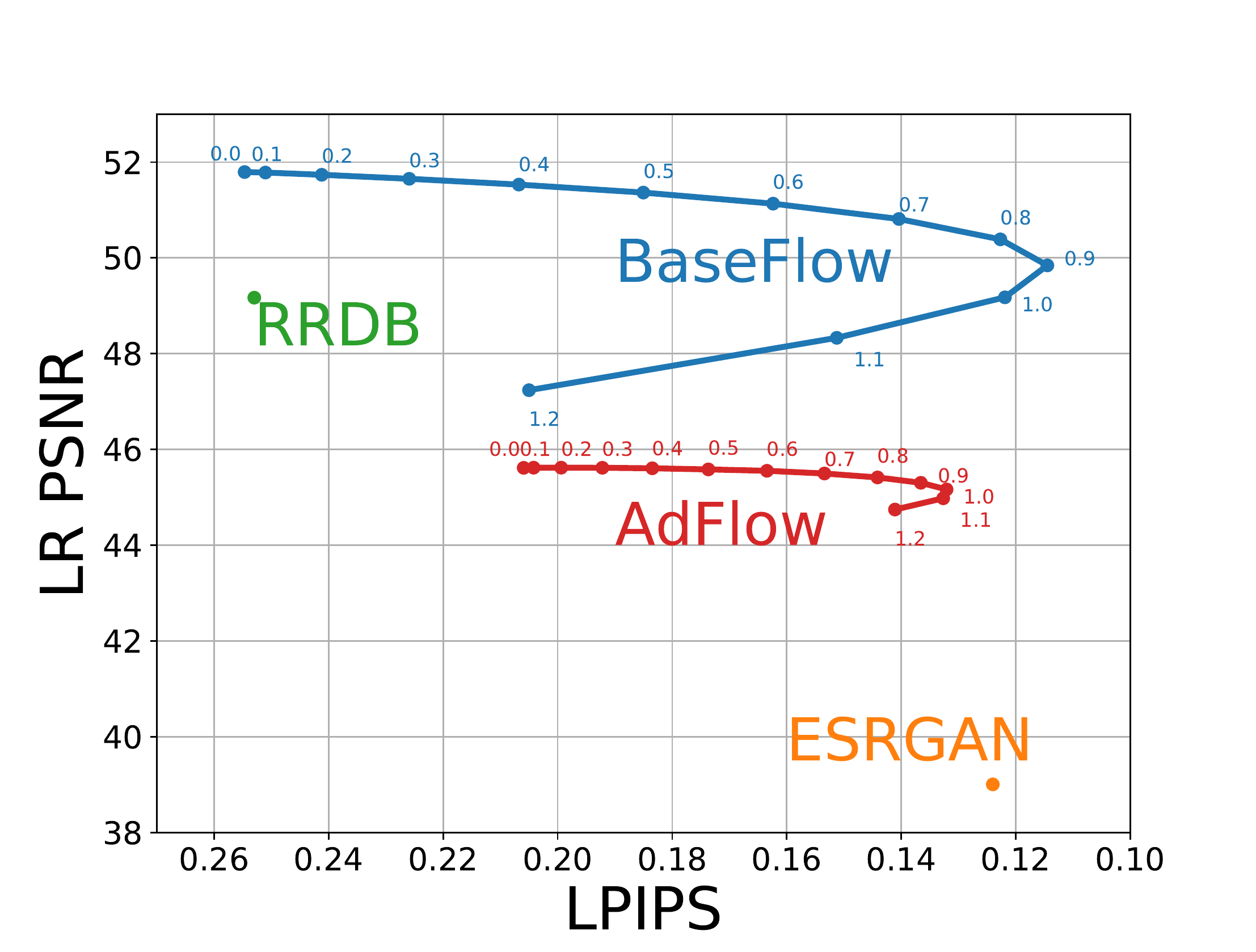}
    \caption{Influence of temperature parameter $\tau$ during inference on perception and fidelity. As opposed to AdFlow, ESRGAN and RRDB have only a single operating point. ($4\times$ super-resolution)}
    \label{fig:PercepDist_4x}\vspace{-5mm}
\end{figure}

\begin{figure}[b]
    \includegraphics[width=\linewidth]{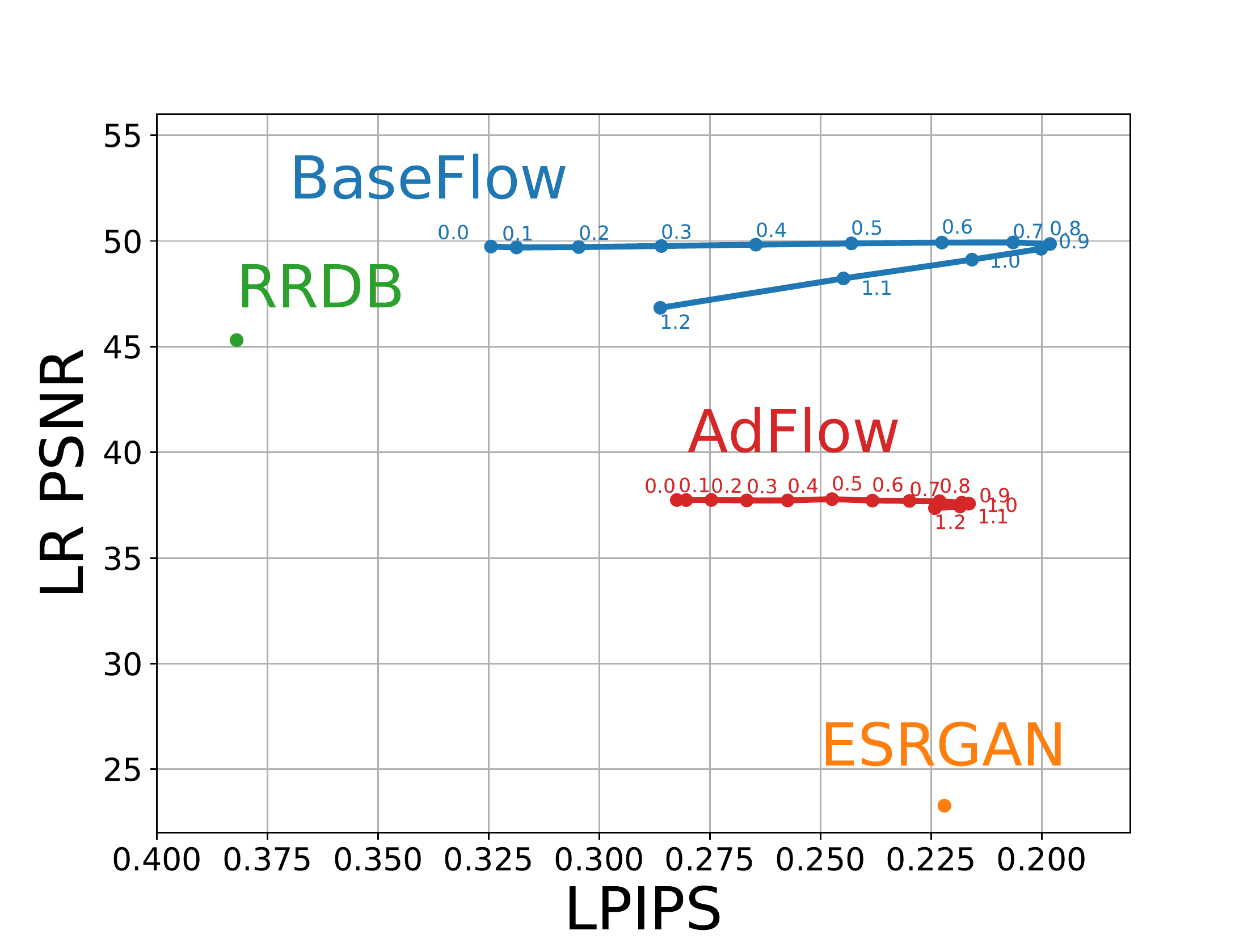}
    \caption{Influence of temperature parameter $\tau$ during inference on perception and fidelity. As opposed to AdFlow, ESRGAN and RRDB have only a single operating point. ($6\times$ super-resolution)}
    \label{fig:PercepDist_6x}\vspace{-3mm}
\end{figure}

\begin{figure}[t]
    \includegraphics[width=\linewidth]{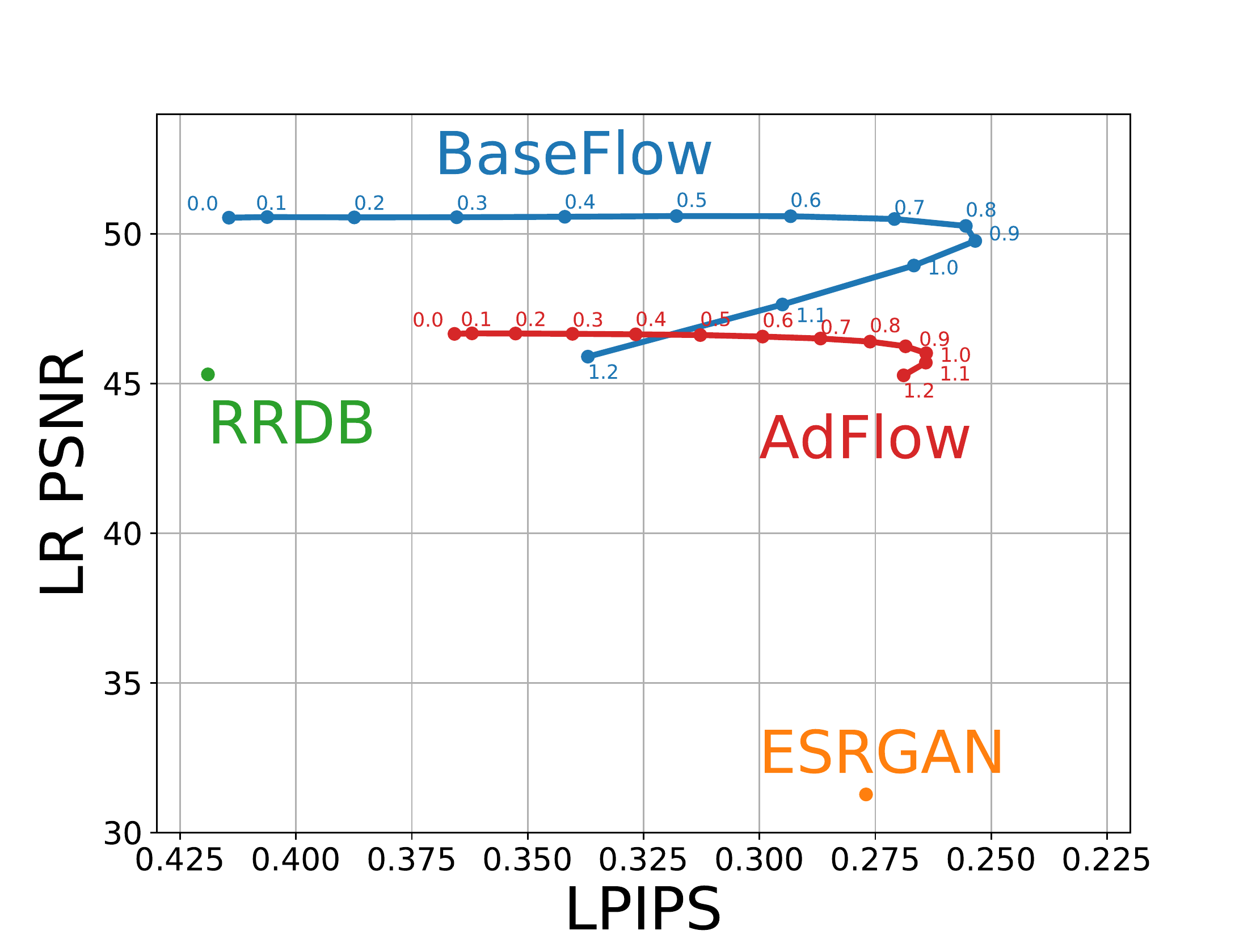}
    \caption{Influence of temperature parameter $\tau$ during inference on perception and fidelity. As opposed to AdFlow, ESRGAN and RRDB have only a single operating point. ($8\times$ super-resolution)}
    \label{fig:PercepDist_8x}\vspace{0mm}
\end{figure}

\section{User Study}
\label{sec:userStudy}

\begin{figure*}[t]
    \centering
    \includegraphics[width=\linewidth]{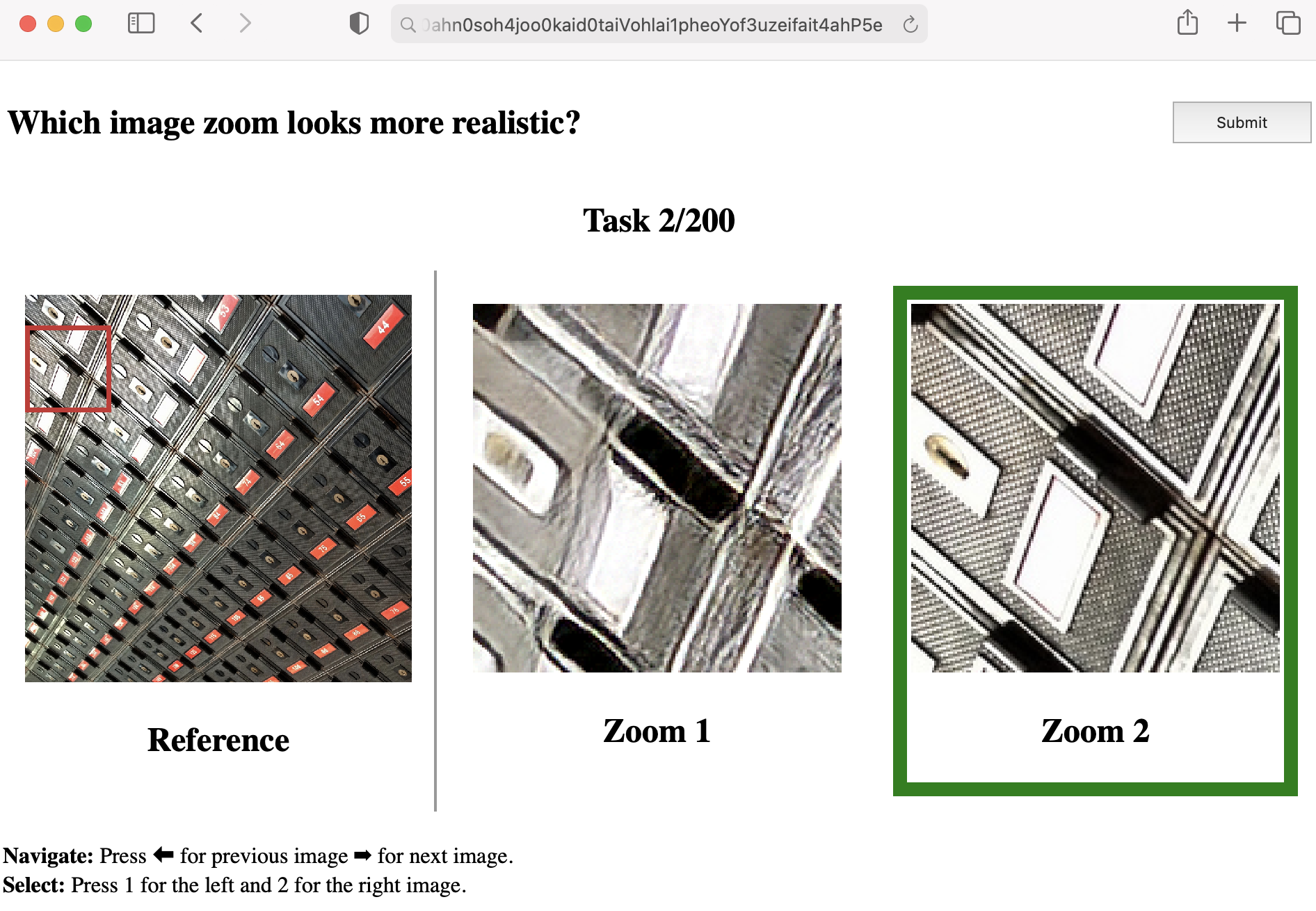}
    \caption{Screenshot of the web GUI of our user study. The `Reference' provides an overview and indicates which part of an image should be considered. The images `Zoom 1' and `Zoom 2' are the two candidates, where the latter was selected to look more realistic.}\vspace{0mm}%
    \label{fig:mturk_gui}
\end{figure*}

As described in Sec.~4.1 in the main paper, we conduct the user study. The GUI interface is shown in Figure~\ref{fig:mturk_gui}. 
We ask the user to evaluate which image of the two looks more realistic.
To select the chosen image, the user presses the \texttt{1} key for the left and \texttt{2} key for the right images.
Once a selection was made, the user can see the next image using the arrow right key until they have completed all tasks.
Finally, the form is submitted using the button on the top right.

\begin{figure*}[t]
    \newcommand\widthMturk{0.2}
    \newcommand{\MturkFirst}[1]{\includegraphics[trim=38 38 30 33,clip]{supplement/figures_supp/mturk/bars_#1.pdf}}
    \newcommand{\MturkNormal}[1]{\includegraphics[trim=221 38 30 33,clip]{supplement/figures_supp/mturk/bars_#1.pdf}}
    \resizebox{1.00\linewidth}{!}{%
    \MturkFirst{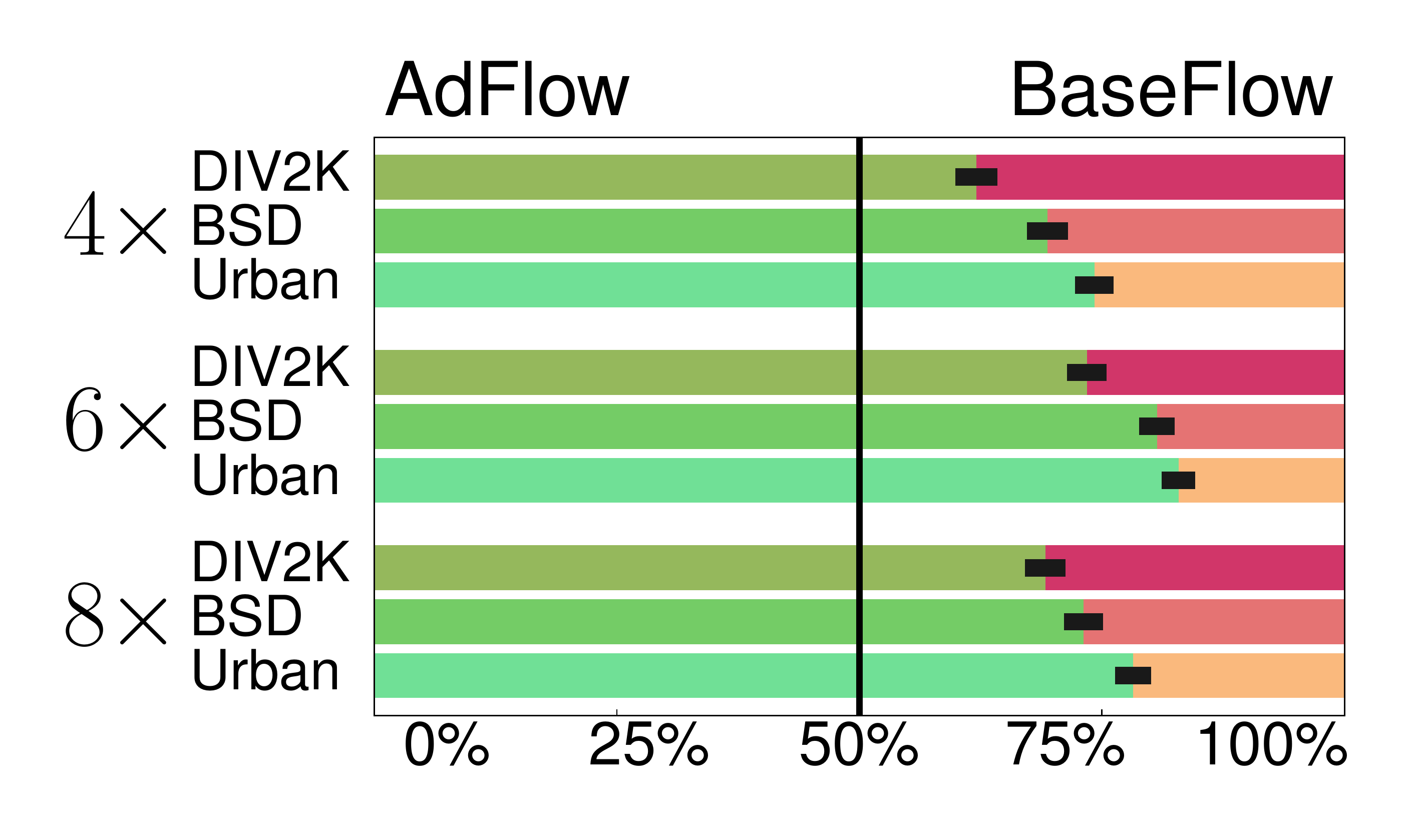}~~~~~~~~~~%
    \MturkNormal{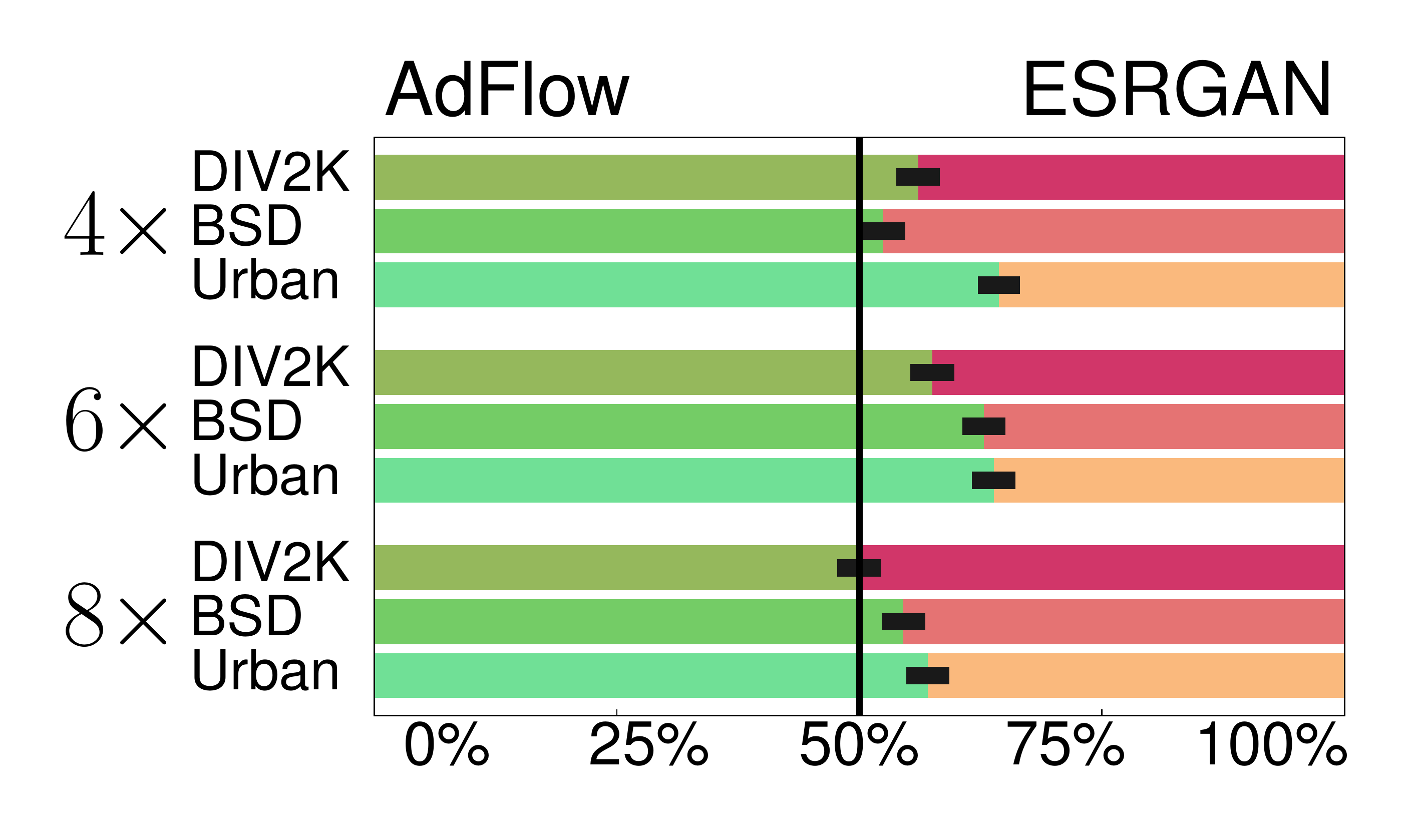}~~~~~~~~~~%
    \MturkNormal{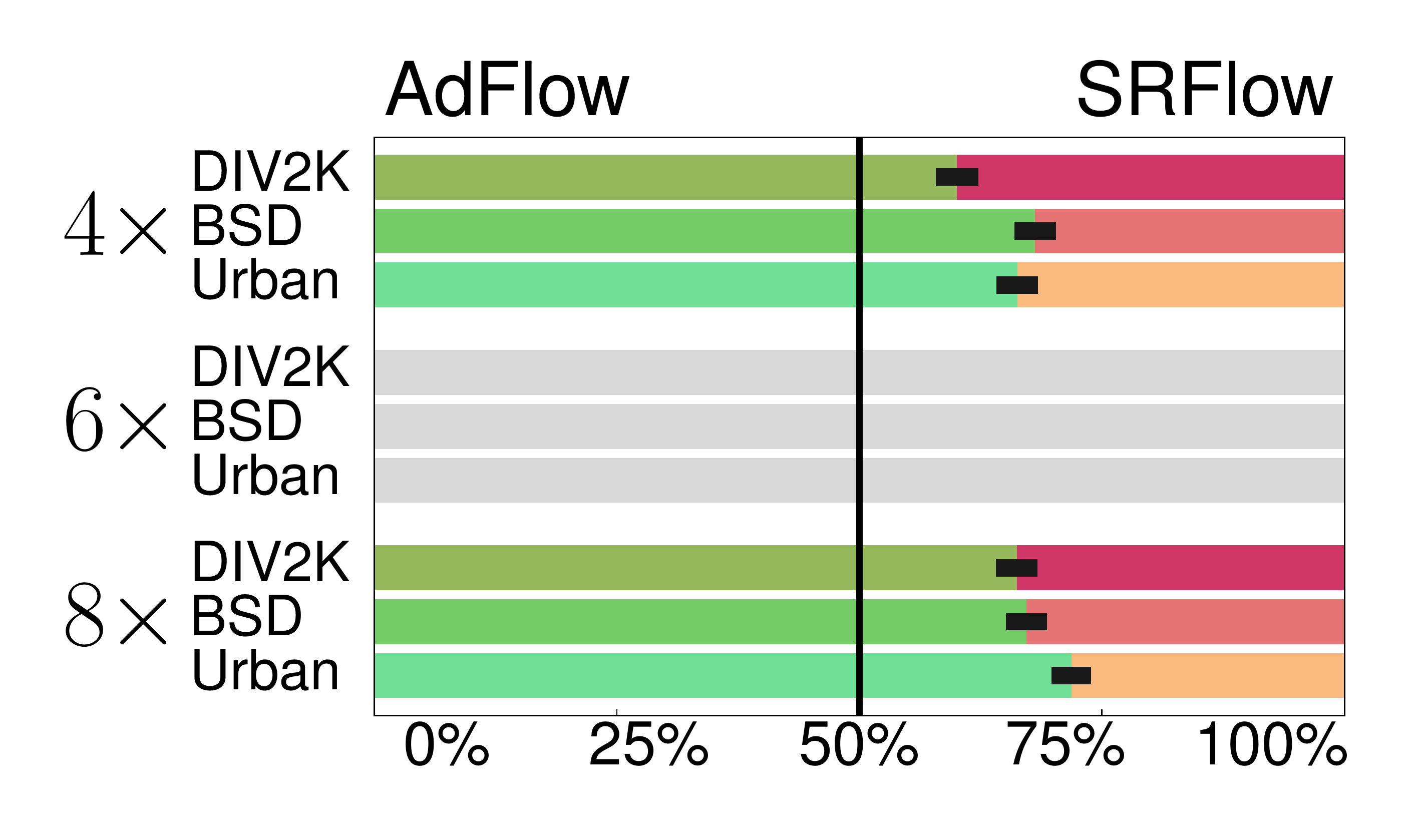}~~~~~~~~~~%
    \MturkNormal{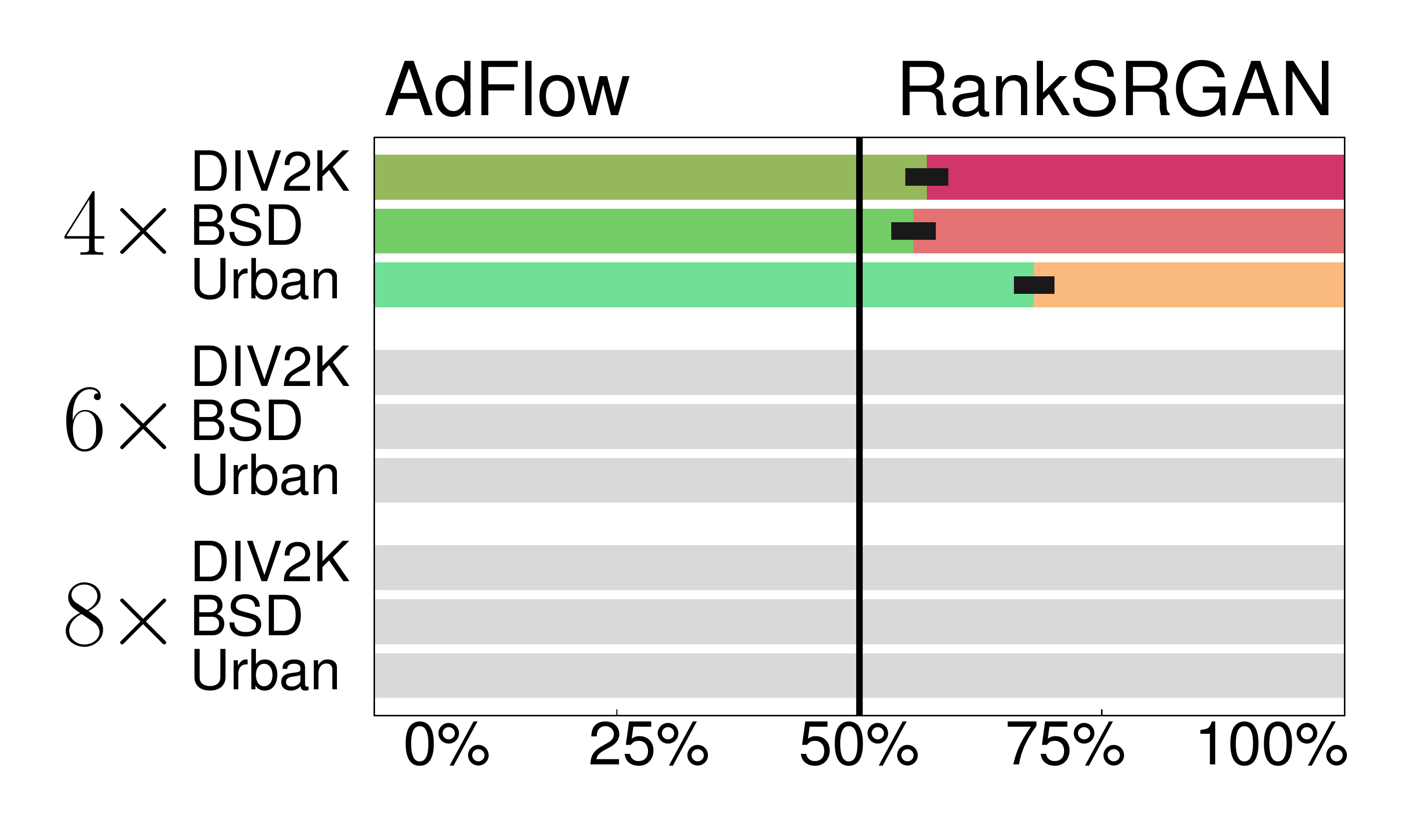}~~~~~~~~~~%
    \MturkNormal{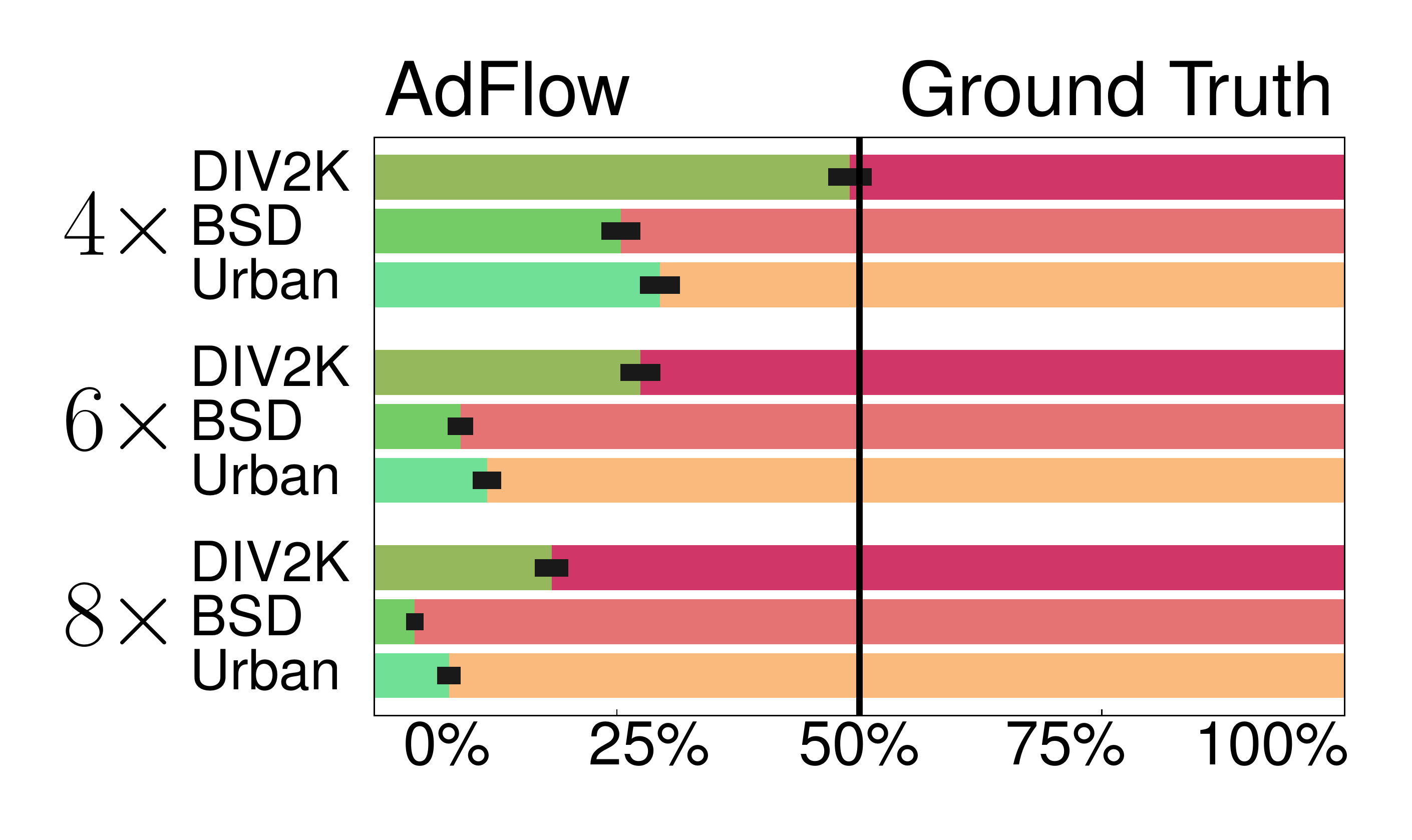}%
    }
    \caption{User study results, as the percentage of votes favoring the photo-realism of AdFlow (green) versus each other method (red). A bar represents 1500 user votes. The 95\% confidence interval is in black. We compare on DIV2K, BSD100, Urban100 for $4\times$, $6\times$, and $8\times$. }
    \label{fig:mturk}
\vspace{0mm}
\end{figure*}

To increase the data quality, we use a filtering mechanism.
For that, we add redundant questions and reject submissions that have a low self-consistency.
A visualization of results of the study results is shown in Fig.~\ref{fig:mturk}.
The green bars display the percentage of votes favoring the photo-realism of AdFlow, while the red bars show the percentage favoring the other method.
We display the statistical significance by showing the 95\% confidence interval in black.
Examples for images used in our study are shown in \texttt{visuals.html}.

\section{Analysis of Sampling Temperature}
\label{sec:percDist}

Here, we analyze the trade-off between the image quality, in terms of LPIPS~\cite{zhang2018unreasonable}, and the consistency to the low-resolution input in terms of LR PSNR when varying the sampling temperature. We sample from the latent space with a Gaussian prior distribution $z \sim \mathcal{N}(0, \tau I)$ with variance $\tau$. The latter is usually termed the sampling temperature \cite{KingmaB14Adam}.
Similar to~\cite{srflow} we can set the operation point by adjusting the temperature $\tau$. 
Figures~\ref{fig:PercepDist_4x},~\ref{fig:PercepDist_6x}~and~\ref{fig:PercepDist_8x} show that ESRGAN~\cite{lim2017EDSR} trades off much more low-resolution consistency to improve the perceptual quality than AdFlow.
The best trade-off is achieved at $\tau=0.9$ for BaseFlow and $\tau=1$ for AdFlow, as used in the main paper.

\section{Minimal $L_1$ generalization}
\label{sec:varL1}
 \begin{figure*}[t]
        \centering%
        \setlength{\lineskip}{0.005\linewidth}
        \newcommand{\img}[1]{%
        \includegraphics[width=0.1203125\linewidth]{supplement/figures_supp/varL1_div2k_8x/#1/Low_Resolution}\hspace{0.005\linewidth}%
\includegraphics[width=0.1203125\linewidth]{supplement/figures_supp/varL1_div2k_8x/#1/RRDB_cite_wang2018esrgan_}\hspace{0.005\linewidth}%
\includegraphics[width=0.1203125\linewidth]{supplement/figures_supp/varL1_div2k_8x/#1/RRDB_cite_wang2018esrgan_Adaptive_Variance}\hspace{0.005\linewidth}%
\includegraphics[width=0.1203125\linewidth]{supplement/figures_supp/varL1_div2k_8x/#1/ESRGAN_cite_wang2018esrgan_}\hspace{0.005\linewidth}%
\includegraphics[width=0.1203125\linewidth]{supplement/figures_supp/varL1_div2k_8x/#1/ESRGAN_cite_wang2018esrgan_Adaptive_Variance}\hspace{0.005\linewidth}%
\includegraphics[width=0.1203125\linewidth]{supplement/figures_supp/varL1_div2k_8x/#1/BaseFlow_text_RRDB_}\hspace{0.005\linewidth}%
\includegraphics[width=0.1203125\linewidth]{supplement/figures_supp/varL1_div2k_8x/#1/AdFlow_text_RRDB_}\hspace{0.005\linewidth}%
\includegraphics[width=0.1203125\linewidth]{supplement/figures_supp/varL1_div2k_8x/#1/Ground_Truth}%
        }%
        \img{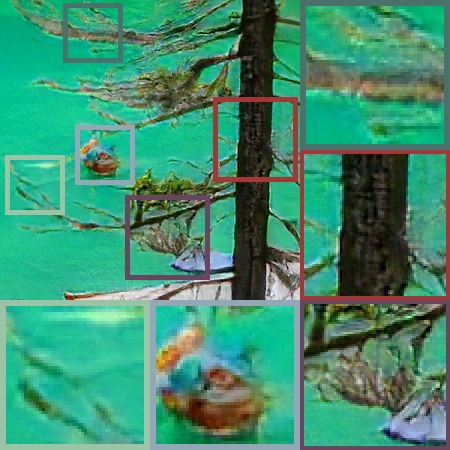}
\img{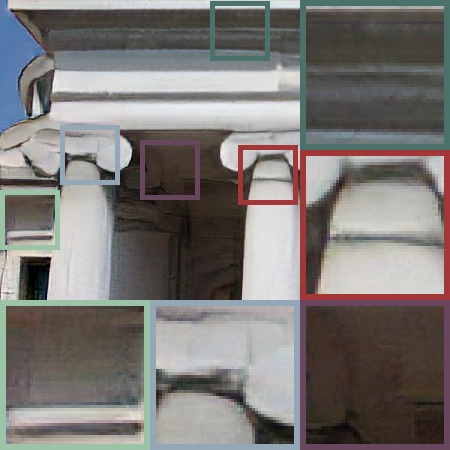}
\img{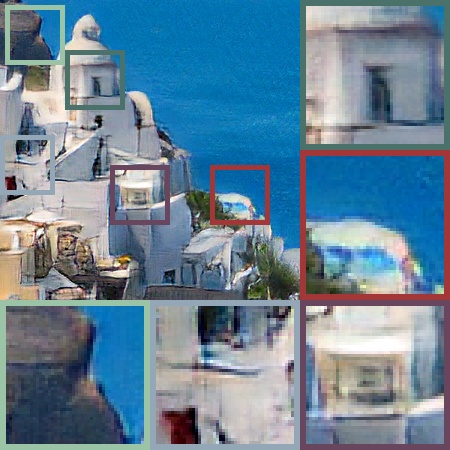}
        \resizebox{1.00\linewidth}{!}{%
        \begin{tabular}{ C{2.8cm} C{2.8cm} C{2.8cm} C{2.8cm} C{2.8cm} C{2.8cm} C{2.8cm} C{2.8cm} }
Low Resolution & RRDB~\cite{wang2018esrgan} & RRDB~\cite{wang2018esrgan} + Adaptive Variance & ESRGAN~\cite{wang2018esrgan} & ESRGAN~\cite{wang2018esrgan} + Adaptive Variance & BaseFlow & AdFlow & Ground Truth
\end{tabular}
        }

     \caption{Qualitative comparison of standard approach and $L_1$ with learned variance on the DIV2K~\cite{div2k} validation set. ($8\times$)}
     \label{fig:varL1_div2k_8x}
 \end{figure*}

Here we provide further theoretical and empirical analysis when generalizing the $L_1$ loss with normalizing flows.

\subsection{Relation of Flow loss to $L_1$}
We here derive the generalized $L_1$ objective (Eq.~(4) in the main paper) from the Normalizing Flow formulation as a 1-layer special case.
Let $z\!\sim\! p_z(z) \!=\! \frac{1}{2^D} e^{-\left\|z\right\|_1}$ be standard Laplace.
We use the function $f$ defined as (Eq.~(3) of the main paper), 
\begin{equation}
\label{eq:scaleL1-f}
    z = f(y;x) = \frac{y - g(x)}{b(x)} \,.
\end{equation}
we obtain the inverse as, 
\begin{equation}
    y = f^{-1}(z;x) = b(x) \cdot z + g(x)
\end{equation}
Since $z\!\sim\! p_z(z)$ is a standard Laplace distribution, it is easy to see that $y \!\sim\! p(y|x; \theta) = \mathcal{L}(y; g(x), b(x))$ as given by Eq.~(4) in the main paper, that is 
\begin{equation}
\label{eq:scaleL1}
-\log p(y|x; \theta) \propto \left\|\frac{y - g(x)}{b(x)}\right\|_1 +  \sum_{ijc} \log b(x)_{ijc} \,.
\end{equation}
Hence, \eqref{eq:scaleL1-f} is the flow $f$ of \eqref{eq:scaleL1}.
Inserting \eqref{eq:scaleL1-f} into the NLL formula for flows (Eq.~(6a) in the main paper) gives,
\begin{align}
&-\log p(\by | \bx, \bT) \!=\! 
    -\log p(z) \!- \!\log \left| \det \frac{\partial\fT}{\partial\by}(\by; \bx) \right| \nonumber\\
&= -\log \frac{1}{2^D} e^{-\|\frac{y-g(x)}{b(x)}\|_1} - \log | \det \text{diag}(\frac{1}{b(x)})| \nonumber\\
&= D\log 2 + \left\|\frac{y-g(x)}{b(x)}\right\|_1 - \log |\prod_{ijc} \frac{1}{b(x)_{ijc}}| \nonumber\\
&\propto \left\|\frac{y-g(x)}{b(x)}\right\|_1 + \sum_{ijc} \log b(x)_{ijc} \, .
\end{align}
Here, the Jacobian $\frac{\partial\fT}{\partial\by} = \text{diag}(\frac{1}{b(x)})$ is a diagonal matrix with elements $\frac{1}{b(x)_{ijk}}$. The final result thus corresponds to the NLL derived directly in \eqref{eq:scaleL1}. We therefore conclude that the generalized $L_1$ objective is a special case given by the 1-layer normalizing flow defined in \eqref{eq:scaleL1-f}.

\subsection{Empirical Analysis}

We report results for the intermediate step of predicting an adaptive variance according to the Laplacian model described in Section 3.1, Equations~(3)-(4) of the main paper.
Those three channels predict the log-scale $a(x) = \log(b(x))$ of the Laplace distribution. The loss in Eq.~(4) of the main paper can thus be written as,
\begin{equation}
    -\log p(y|x; \theta) \propto \left\|\frac{y - g(x)}{\exp(a(x))}\right\|_1 +  \sum_{ijc} a(x)_{ijc} \,.
\end{equation}

We notice that even this extension of the $L_1$ objective reduces the conflict with the adversarial loss to some extent.
The effective removal of artifacts for $8\times$ super-resolution is especially apparent in the first row of Figure~\ref{fig:varL1_div2k_8x} between ESRGAN~\cite{wang2018esrgan} and ESRGAN + Adaptive Variance.
Our further generalization of $L_1$ loss continues to improve the quality of the super-resolutions.

As the increase in visual quality alone would not be a good indicator for a reduced conflict of objectives,
we also report the low-resolution consistency in Table~\ref{tab:varL1_div2k_8x} which improves by $2.91$dB from ESRGAN~\cite{wang2018esrgan} to ESRGAN + Adaptive Variance.
An additional generalization to BaseFlow and AdFlow leads to a further improved low-resolution consistency.
Based on observing an improved visual quality and low-resolution consistency, we conclude that the minimally generalized $L_1$ loss reduces the conflict in objectives, which further validates our strategy of replacing the $L_1$ with a more flexible generalization.

\begin{table}[t]
    \centering%
    \resizebox{\linewidth}{!}{%

\begin{tabular}{lllll}
\toprule
                                                  & PSNR~$\uparrow$ & SSIM~$\uparrow$ & LPIPS~$\downarrow$ & LR-PSNR~$\uparrow$ \\
\midrule
                       RRDB~\cite{wang2018esrgan} &           25.52 &           0.697 &              0.419 &              45.31 \\
   RRDB~\cite{wang2018esrgan} + Adaptive Variance &           25.47 &           0.696 &              0.418 &              44.51 \\
                     ESRGAN~\cite{wang2018esrgan} &           22.14 &           0.578 &              0.277 &              31.28 \\
 ESRGAN~\cite{wang2018esrgan} + Adaptive Variance &           22.94 &           0.593 &              0.280 &              34.19 \\
                         BaseFlow$_{\text{RRDB}}$ &           23.58 &           0.595 &              0.253 &              49.78 \\
                           AdFlow$_{\text{RRDB}}$ &           23.45 &           0.602 &              0.253 &              47.54 \\
\bottomrule
\end{tabular}

    }
    \caption{Quantitative results of standard approach and $L_1$ with learned variance on the DIV2K~\cite{div2k} validation set. ``Adaptive variance'' indicates the generalized $L_1$ loss with predicted variance, as described in Sec.~\ref{sec:varL1}. ($8\times$)}%
    \label{tab:varL1_div2k_8x}
    \vspace{0mm}
\end{table}

\section{Detailed Results}
\label{sec:extendedEval}

In this section, we provide an extended quantitative and qualitative analysis of the same BaseFlow and AdFlow networks evaluated in the main paper.
For completeness, we here provide the PSNR, SSIM and LPIPS on the DIV2K, BSD100, and Urban100 datasets.
Results are reported in Tables~\ref{tab:sota_div2k_4x},~\ref{tab:sota_div2k_6x}~and~\ref{tab:sota_div2k_8x}. However, note that these metrics do not well reflect photo-realism, as discussed in Sec.~4.1 in the main paper.

Further qualitative results for the scale levels $4\times$,~$6\times$~and~$8\times$ are provided in Figures~\ref{fig:sota_x4},~\ref{fig:sota_x6}~and~\ref{fig:sota_x8} respectively.

\newcommand{\computeMetrics}[1]{Approximations for perceptual quality on the sets DIV2K (val.), BSD100, and Urban100 ($#1\times$). Since~\cite{ledig2017photo,Sajjadi17EnhanceNet,wang2018esrgan,lugmayrICCVW2019,fritsche19FrequencySeparation,Ji2020Impressionism,srflow} showed the limitations of calculated metrics for SR our main metric is the human study.}

\begin{table}[t]
    \centering%
    \resizebox{0.95\linewidth}{!}{%
    \begin{tabular}{lllllll}
    \toprule
    &                                     & PSNR~$\uparrow$ & SSIM~$\uparrow$ & LPIPS~$\downarrow$ & LR-PSNR~$\uparrow$ \\
    \midrule
    \multirow{7}{1mm}{\resizebox{2mm}{!}{\rotatebox{90}{DIV2K}}}
    & Bicubic                             & 26.69           & 0.766           & 0.409              & 38.69              \\
    & RRDB~\cite{wang2018esrgan}          & 29.44           & 0.844           & 0.253              & 49.17              \\
    & ESRGAN~\cite{wang2018esrgan}        & 26.20           & 0.747           & 0.124              & 39.01              \\
    & RankSRGAN~\cite{zhang2019ranksrgan} & 26.55           & 0.750           & 0.128              & 42.33              \\
    & SRFlow~\cite{srflow}                & 27.08           & 0.756           & 0.120              & 49.97              \\
    & BaseFlow            & 27.21           & 0.760           & 0.118              & 49.88              \\
    & AdFlow              & 27.02           & 0.768           & 0.132              & 45.17              \\
    \midrule
    \multirow{7}{1mm}{\resizebox{2mm}{!}{\rotatebox{90}{BSD100}}}
    & Bicubic                             & 22.40           & 0.508           & 0.713              & 37.13              \\
    & RRDB~\cite{wang2018esrgan}          & 23.58           & 0.572           & 0.554              & 45.26              \\
    & ESRGAN~\cite{wang2018esrgan}        & 20.99           & 0.462           & 0.332              & 31.68              \\
    & SRFlow~\cite{srflow}                & 21.76           & 0.467           & 0.335              & 51.01              \\
    & BaseFlow            & 22.03           & 0.478           & 0.325              & 50.17              \\
    & AdFlow              & 22.01           & 0.486           & 0.327              & 48.78              \\
    \midrule
    \multirow{7}{1mm}{\resizebox{2mm}{!}{\rotatebox{90}{Urban100}}}
    & Bicubic                             & 19.31           & 0.477           & 0.686              & 33.93              \\
    & RRDB~\cite{wang2018esrgan}          & 21.15           & 0.603           & 0.401              & 43.33              \\
    & ESRGAN~\cite{wang2018esrgan}        & 18.43           & 0.475           & 0.306              & 28.88              \\
    & SRFlow~\cite{srflow}                & 19.29           & 0.501           & 0.309              & 48.11              \\
    & BaseFlow            & 19.72           & 0.513           & 0.304              & 48.71              \\
    & AdFlow              & 19.04           & 0.506           & 0.278              & 44.67              \\
    \bottomrule
\end{tabular}

    }
    \caption{\computeMetrics{4}}\vspace{0mm}%
    \label{tab:sota_div2k_4x}
    \vspace{0mm}
\end{table}

\begin{table}[t]
    \centering%
    \resizebox{0.95\linewidth}{!}{%
    \begin{tabular}{lllllll}
    \toprule
    &                              & PSNR~$\uparrow$ & SSIM~$\uparrow$ & LPIPS~$\downarrow$ & LR-PSNR~$\uparrow$ \\
    \midrule
    \multirow{4}{1mm}{\resizebox{2mm}{!}{\rotatebox{90}{DIV2K}}}
    & Bicubic                      & 24.87           & 0.680           & 0.519              & 37.78              \\
    & RRDB                         & 26.51           & 0.741           & 0.382              & 46.86              \\
    & ESRGAN~\cite{wang2018esrgan} & 23.16           & 0.629           & 0.222              & 33.21              \\
    & BaseFlow     & 24.04           & 0.621           & 0.216              & 49.12              \\
    & AdFlow       & 23.94           & 0.6505          & 0.216              & 37.57              \\
    \midrule
    \multirow{4}{1mm}{\resizebox{2mm}{!}{\rotatebox{90}{BSD100}}}
    & Bicubic                      & 23.26           & 0.564           & 0.645              & 37.51              \\
    & RRDB                         & 24.42           & 0.625           & 0.507              & 46.41              \\
    & ESRGAN~\cite{wang2018esrgan} & 21.42           & 0.501           & 0.288              & 32.76              \\
    & BaseFlow     & 21.98           & 0.500           & 0.274              & 48.60              \\
    & AdFlow       & 22.17           & 0.533           & 0.269              & 38.52              \\
    \midrule
    \multirow{4}{1mm}{\resizebox{2mm}{!}{\rotatebox{90}{Urban100}}}
    & Bicubic                      & 20.20           & 0.541           & 0.606              & 34.24              \\
    & RRDB                         & 21.95           & 0.650           & 0.371              & 44.81              \\
    & ESRGAN~\cite{wang2018esrgan} & 19.43           & 0.541           & 0.251              & 30.87              \\
    & BaseFlow     & 20.43           & 0.564           & 0.255              & 47.90              \\
    & AdFlow       & 20.26           & 0.583           & 0.235              & 36.01              \\
    \bottomrule
\end{tabular}
    }
    \caption{\computeMetrics{6}}\vspace{0mm}%
    \label{tab:sota_div2k_6x}
    \vspace{0mm}
\end{table}

\begin{table}[t]
    \centering%
    \resizebox{0.95\linewidth}{!}{%
    \begin{tabular}{lllllll}
    \toprule
    &                              & PSNR~$\uparrow$ & SSIM~$\uparrow$ & LPIPS~$\downarrow$ & LR-PSNR~$\uparrow$ \\
    \midrule
    \multirow{7}{1mm}{\resizebox{2mm}{!}{\rotatebox{90}{DIV2K}}}
    & Bicubic                      & 23.74           & 0.627           & 0.584              & 37.14              \\
    & RRDB~\cite{wang2018esrgan}   & 25.52           & 0.697           & 0.419              & 45.31              \\
    & ESRGAN~\cite{wang2018esrgan} & 22.14           & 0.578           & 0.277              & 31.28              \\
    & SRFlow~\cite{srflow}         & 23.04           & 0.578           & 0.275              & 49.02              \\
    & BaseFlow     & 23.58           & 0.595           & 0.253              & 49.78              \\
    & AdFlow       & 23.38           & 0.600           & 0.264              & 46.02              \\
    \midrule
    \multirow{7}{1mm}{\resizebox{2mm}{!}{\rotatebox{90}{BSD100}}}
    & Bicubic                      & 22.40           & 0.508           & 0.713              & 37.13              \\
    & RRDB~\cite{wang2018esrgan}   & 23.58           & 0.572           & 0.554              & 45.26              \\
    & ESRGAN~\cite{wang2018esrgan} & 20.99           & 0.462           & 0.332              & 31.68              \\
    & SRFlow~\cite{srflow}         & 21.76           & 0.467           & 0.335              & 51.01              \\
    & BaseFlow     & 22.03           & 0.478           & 0.325              & 50.17              \\
    & AdFlow       & 22.01           & 0.486           & 0.327              & 48.78              \\
    \midrule
    \multirow{7}{1mm}{\resizebox{2mm}{!}{\rotatebox{90}{Urban100}}}
    & Bicubic                      & 19.31           & 0.477           & 0.686              & 33.93              \\
    & RRDB~\cite{wang2018esrgan}   & 21.15           & 0.603           & 0.401              & 43.33              \\
    & ESRGAN~\cite{wang2018esrgan} & 18.43           & 0.475           & 0.306              & 28.88              \\
    & SRFlow~\cite{srflow}         & 19.29           & 0.501           & 0.309              & 48.11              \\
    & BaseFlow     & 19.72           & 0.513           & 0.304              & 48.71              \\
    & AdFlow       & 19.04           & 0.506           & 0.278              & 44.67              \\
    \bottomrule
\end{tabular}
    }
    \caption{\computeMetrics{8}}\vspace{0mm}%
    \label{tab:sota_div2k_8x}
    \vspace{0mm}
\end{table}

\begin{figure*}[t]
\rotatebox[origin=c]{90}{\resizebox{7.2cm}{!}{
		\begin{tabular}{ C{3cm} C{3cm} C{3cm}}
			Urban100 & BSD100 & DIV2K
		\end{tabular}
}}\vspace{0mm}%
\begin{minipage}{0.95\linewidth}%
\includegraphics[width=\linewidth]{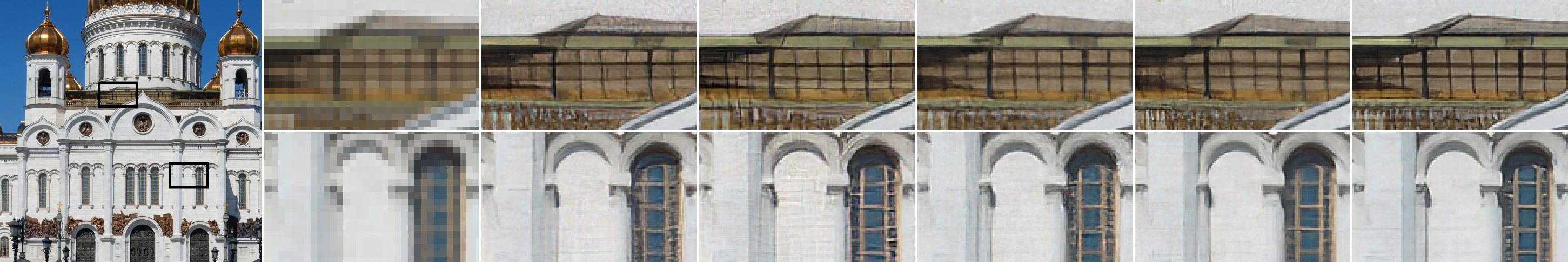}
\includegraphics[width=\linewidth]{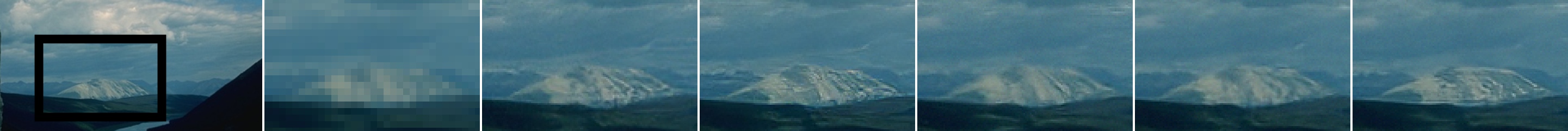}
\includegraphics[width=\linewidth]{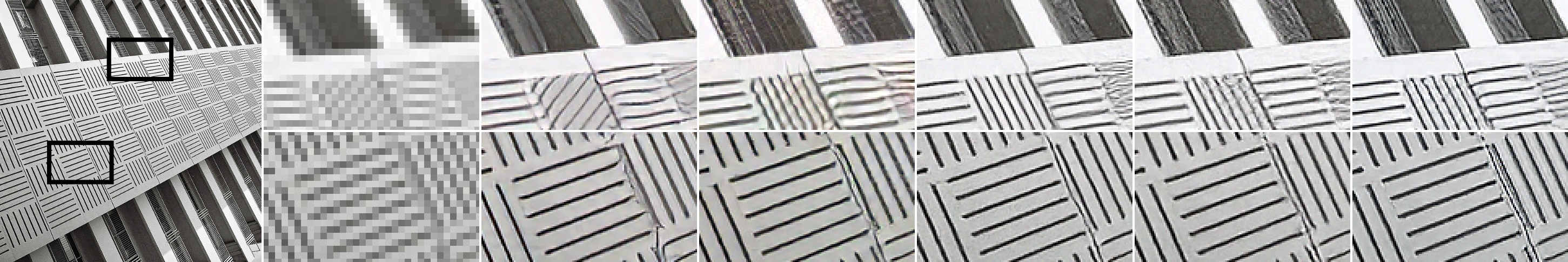}
\end{minipage}
\resizebox{1.00\linewidth}{!}{%
\begin{tabular}{ C{4.2cm} C{3cm} C{3cm} C{3cm} C{3cm} C{3cm} C{3cm} C{0.5cm} }
& Low Resolution & RankSRGAN & ESRGAN\cite{wang2018esrgan} & SRFlow\cite{srflow} & BaseFlow & AdFlow &
\end{tabular}}\vspace{-1mm}
\caption{Qualitative comparison with state-of-the-art approaches on the DIV2K (val), BSD100 and Urban100 set for $4\times$ SR.}
\vspace{-1mm}
\label{fig:sota_x4}
\end{figure*}

\begin{figure*}[t]
\rotatebox[origin=c]{90}{\resizebox{8.7cm}{!}{
		\begin{tabular}{ C{3cm} C{3cm} C{3cm}}
			Urban100 & BSD100 & DIV2K
		\end{tabular}
}}\vspace{0mm}%
\begin{minipage}{0.95\linewidth}
\includegraphics[width=\linewidth]{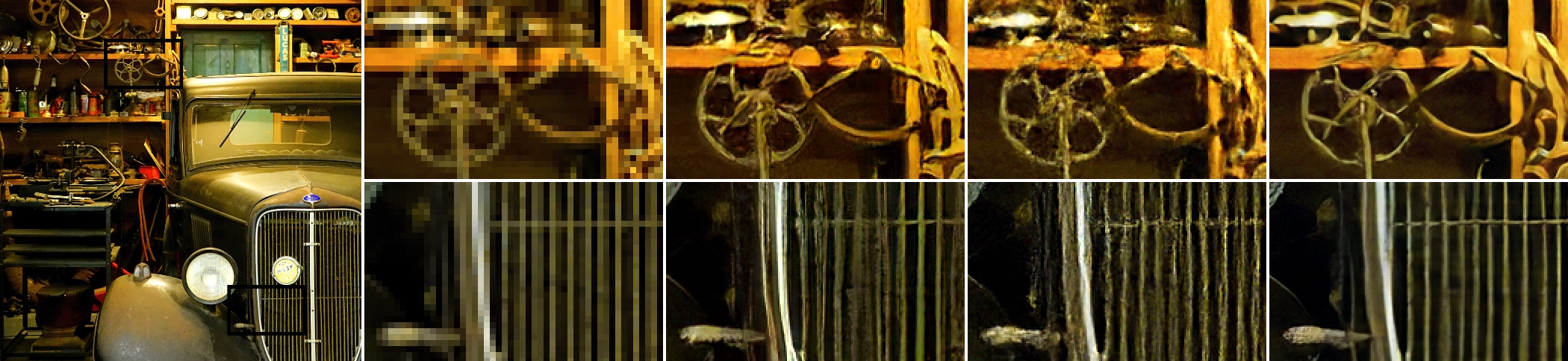}
\includegraphics[width=\linewidth]{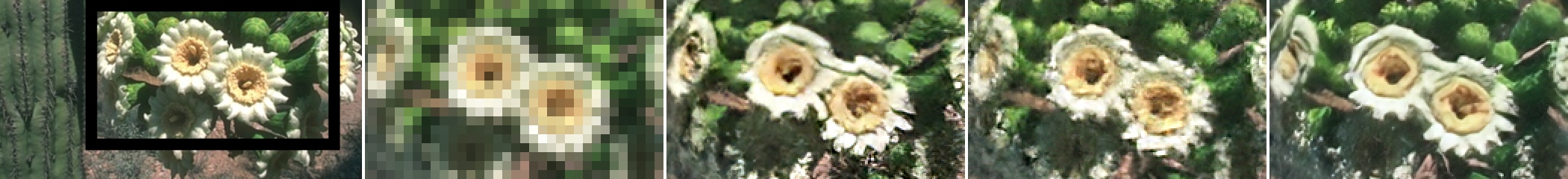}
\includegraphics[width=\linewidth]{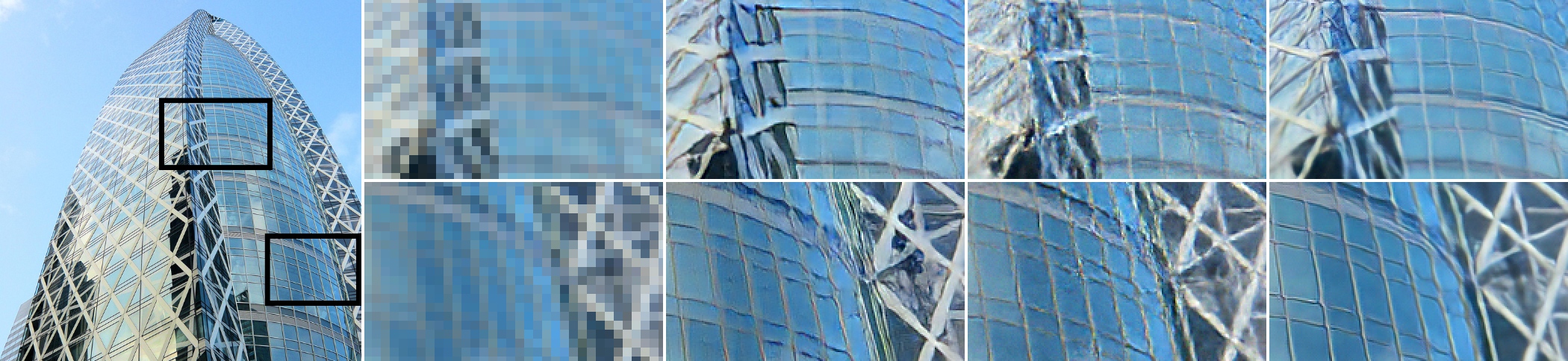}
\end{minipage}
\resizebox{1.00\linewidth}{!}{%
\begin{tabular}{ C{4.2cm} C{3cm} C{3cm} C{3cm} C{3cm} C{0.15cm} }
& Low Resolution & ESRGAN\cite{wang2018esrgan} & BaseFlow & AdFlow &
\end{tabular}}\vspace{-1mm}
\caption{Qualitative comparison with state-of-the-art approaches on the DIV2K (val), BSD100 and Urban100 set for $6\times$ SR.}
\label{fig:sota_x6}
\vspace{-3mm}
\end{figure*}

\begin{figure*}[t]
\rotatebox[origin=c]{90}{\resizebox{7cm}{!}{
		\begin{tabular}{ C{3cm} C{3cm} C{3cm}}
			Urban100 & BSD100 & DIV2K
		\end{tabular}
}}\vspace{0mm}%
\begin{minipage}{0.95\linewidth}
\includegraphics[width=\linewidth]{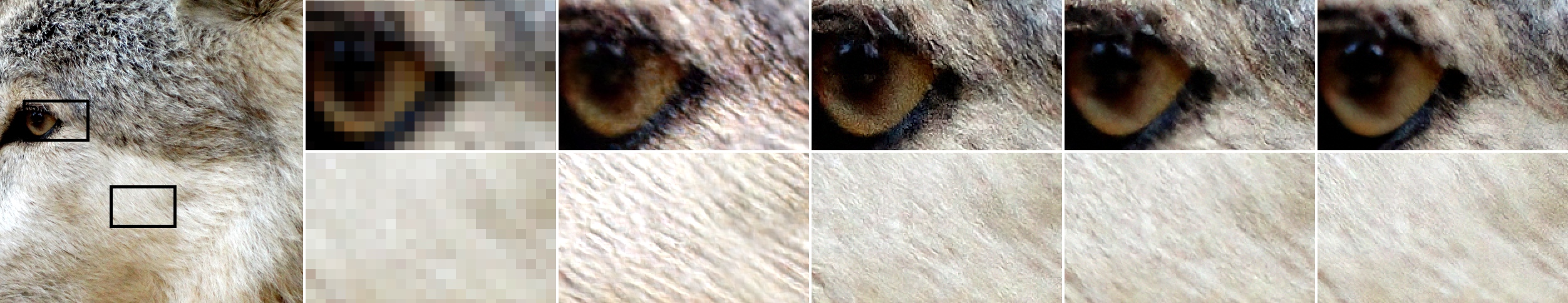}
\includegraphics[width=\linewidth]{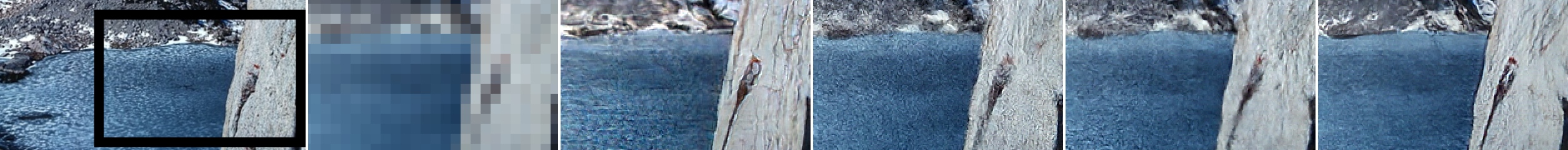}
\includegraphics[width=\linewidth]{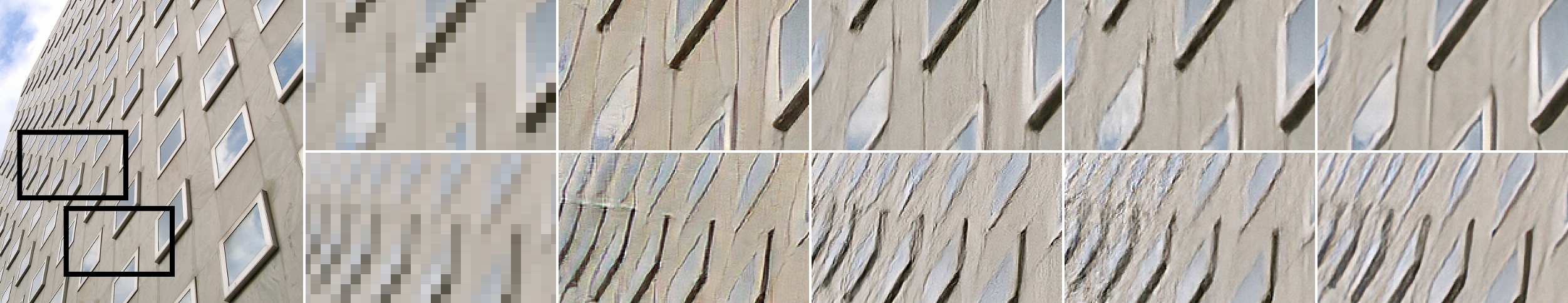}
\end{minipage}
\resizebox{1.00\linewidth}{!}{%
\begin{tabular}{ C{3.9cm} C{3cm} C{3cm} C{3cm} C{3cm} C{3cm} C{0.2cm} }
& Low Resolution & ESRGAN\cite{wang2018esrgan} & SRFlow\cite{srflow} & BaseFlow & AdFlow &
\end{tabular}}\vspace{-1mm}
    \caption{Qualitative comparison with state-of-the-art approaches on the DIV2K (val), BSD100 and Urban100 set for $8\times$ SR.}
\label{fig:sota_x8}
\vspace{-2mm}
\end{figure*}

\end{document}